\documentclass[sigconf]{acmart}

\AtBeginDocument{%
  \providecommand\BibTeX{{%
    \normalfont B\kern-0.5em{\scshape i\kern-0.25em b}\kern-0.8em\TeX}}}

\usepackage{amsmath,amsfonts,bm}

\def\sref#1{\S~\ref{#1}}
\def\eqref#1{equation~\ref{#1}}
\def\Eqref#1{Eq.(\ref{#1})}
\def\1{\bm{1}}

\def\va{{\bm{a}}}
\def\vb{{\bm{b}}}

\def\vh{{\bm{h}}}

\def\vp{{\bm{p}}}

\def\vx{{\textbf{x}}}

\def\mD{{\bm{D}}}

\def\mP{{\bm{P}}}

\def\mW{{\bm{W}}}
\def\mX{{\bm{X}}}

\DeclareMathAlphabet{\mathsfit}{\encodingdefault}{\sfdefault}{m}{sl}
\SetMathAlphabet{\mathsfit}{bold}{\encodingdefault}{\sfdefault}{bx}{n}

\usepackage{enumitem}
\usepackage{hyperref}
\usepackage[capitalize,noabbrev]{cleveref}
\usepackage{booktabs} % For formal tables
\usepackage{graphicx}
\usepackage{nicefrac}
\usepackage{subfigure}
\usepackage{dsfont}
\usepackage{boldline}
\usepackage{bm}
\usepackage{url}
\usepackage{bigstrut}
\usepackage[ruled,vlined]{algorithm2e}
\usepackage{multirow}
\usepackage{mathtools, nccmath}
\usepackage{balance}

\newcommand{\stitle}[1]{\vspace{2mm} \noindent {\bf #1}}
\newcommand{\method}[1]{\textsf{#1}}
\newcommand{\model}{\method{ULTRA-DP}{}}
\newcommand{\reach}[3]{\textsc{Rch}({#1}, {#2}, {#3})}

\usepackage{CJKutf8}

\begin{document}
% \fancyhead{}

%%
%% The "title" command has an optional parameter,
%% allowing the author to define a "short title" to be used in page headers.
\title{ULTRA-DP: Unifying Graph Pre-training with\\ Multi-task Graph Dual Prompt}
\renewcommand{\shorttitle}{ULTRA-DP: Unifying Graph Pre-training with Multi-task Graph Dual Prompt}

\author{Mouxiang Chen}
\affiliation{%
  \institution{Zhejiang University}
  \country{}
}
\email{chenmx@zju.edu.cn}

\author{Zemin Liu}
\authornote{
    Corresponding authors.
}
\affiliation{%
  \institution{National University of Singapore}
  \country{}
}
\email{zeminliu@nus.edu.sg}

\author{Chenghao Liu}
\authornotemark[1]
\affiliation{
  \institution{Salesforce Research Asia}
  \country{}
}
\email{chenghao.liu@salesforce.com}

\author{Jundong Li}
\affiliation{
  \institution{University of Virginia}
  \country{}
}
\email{jundong@virginia.edu}

\author{Qiheng Mao}
\affiliation{
  \institution{Zhejiang University}
  \country{}
}
\email{maoqiheng@zju.edu.cn}

\author{Jianling Sun}
\affiliation{
  \institution{Zhejiang University}
  \country{}
}
\email{sunjl@zju.edu.cn}

%%
%% By default, the full list of authors will be used in the page
%% headers. Often, this list is too long, and will overlap
%% other information printed in the page headers. This command allows
%% the author to define a more concise list
%% of authors' names for this purpose.
% \renewcommand{\shortauthors}{Trovato and Tobin, et al.}
% \renewcommand{\shortauthors}{Mouxiang Chen et al.}

\begin{abstract}

Recent research has demonstrated the efficacy of pre-training graph neural networks (GNNs) to capture the transferable graph semantics and enhance the performance of various downstream tasks. However, the semantic knowledge learned from pretext tasks might be unrelated to the downstream task, leading to a semantic gap that limits the application of graph pre-training. 
To reduce this gap, traditional approaches propose hybrid pre-training to combine various pretext tasks together in a multi-task learning fashion and learn multi-grained knowledge, which, however, cannot distinguish tasks and results in some transferable \textit{task-specific knowledge} distortion by each other. Moreover, most GNNs cannot distinguish nodes located in different parts of the graph, making them fail to learn \textit{position-specific knowledge} and lead to suboptimal performance. In this work, inspired by the \emph{prompt-based tuning} in natural language processing, we propose a unified framework for graph hybrid pre-training which injects the task identification and position identification into GNNs through a prompt mechanism, namely multi-task graph dual prompt (\model). 
Based on this framework, we propose a prompt-based transferability test to find the most relevant pretext task in order to reduce the semantic gap. 
To implement the hybrid pre-training tasks, beyond the classical edge prediction task (node-node level), we further propose a novel pre-training paradigm based on a group of $k$-nearest neighbors (node-group level).
The combination of them across different scales is able to comprehensively express more structural semantics and derive richer multi-grained knowledge.
Extensive experiments show that our proposed \model~can significantly enhance the performance of hybrid pre-training methods and show the generalizability to other pre-training tasks and backbone architectures. \footnote{Code is available at \url{https://github.com/Keytoyze/ULTRA-DP}.}

\end{abstract}

%%
%% The code below is generated by the tool at http://dl.acm.org/ccs.cfm.
%% Please copy and paste the code instead of the example below.
%%


%%
%% Keywords. The author(s) should pick words that accurately describe
%% the work being presented. Separate the keywords with commas.

%%
%% This command processes the author and affiliation and title
%% information and builds the first part of the formatted document.
\maketitle

\newcommand{\eg}{{\it e.g.}}
\newcommand{\etal}{{\it et al.}}
\newcommand{\ie}{{\it i.e.}}

\section{Introduction}

Recent studies have demonstrated the efficacy of using graph neural networks (GNNs) \cite{wu2020comprehensive} to analyze graph data and generate node representations for many real-work tasks, such as node classification \cite{kipf2017semi}, link prediction \cite{zhang2018link}, and recommendation \cite{ying2018graph}. Commonly, training GNNs requires sufficient labeled data, which is often costly and sometimes impossible to obtain \cite{liu2022graph}. To address this issue, graph pre-training is proposed to reduce the dependence on labels \cite{hu2020gpt,lu2021learning,Hu2020Strategies}. By solving a series of pretext tasks from unlabeled data (\eg, edge prediction \cite{hamilton2017inductive}), graph pre-training enables the model to capture the underlying transferable graph knowledge and promote the performance of various downstream tasks even with a limited amount of labeled data.

Despite the success, some gaps exist between pretext tasks and downstream tasks, which limit the application of graph pre-training. For example, edge prediction optimizes binary predictions for node pairs (as a pre-training objective), while node classification (as a downstream task) focuses on multi-class predictions for standalone nodes. The inconsistency of the objectives impedes the knowledge transfer from pre-training on the edge prediction to the downstream node classification task. To alleviate this issue, Sun \etal\ proposed GPPT \cite{sun2022gppt} which reformulates the downstream node classification task as an edge prediction task to bridge the training objective gap. 
However, even if the pre-training and downstream tasks appear to be consistent, there still exists a \textbf{semantic gap} since the underlying knowledge learned from pretext tasks might be irrelevant to the interest of downstream tasks and may even hurt the performance of the latter, leading to negative knowledge transfer \cite{Rosenstein2005ToTO}. 
For example, edge prediction learns the structural semantics of connections between nodes, while node classification requires comprehensive semantics of structure and attributes.
To address this issue, many studies adopt \emph{hybrid training} \cite{hu2020gpt,Hu2020Strategies,manessi2021graph}, which integrates various pretext tasks together in a multi-task learning fashion. This enables GNNs to catch semantic properties of graphs from different views and learn multi-grained knowledge, which further helps to transfer knowledge to downstream tasks.

\begin{figure}[t!]
    \centering
    \includegraphics[width=0.48\textwidth]{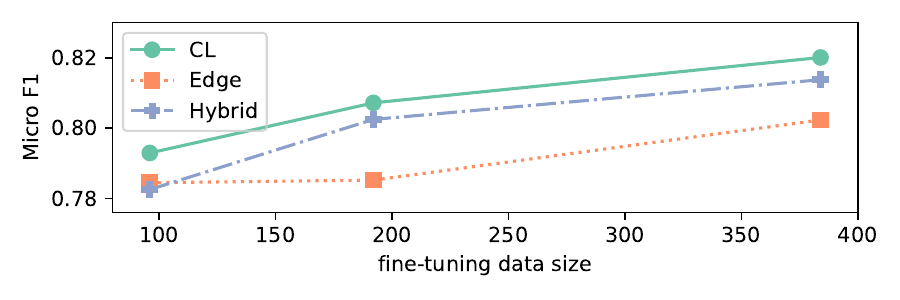}
    \caption{Downstream node classification performance of single and hybrid pre-training tasks. \method{CL}: Contrastive learning, \method{Edge}: Edge prediction. \method{Hybrid}: Combine them together.}
    \label{fig:pilot}
\end{figure}

Though effective, hybrid training is still facing a main challenge in that multiple tasks may share the same GNN parameters and interfere with each other, giving rise to the issue that some useful task-specific knowledge is distorted by other pre-training tasks, which further hurts the fine-tuning performance. For a clear illustration, we carried out a pilot experiment with several prevalent pre-training approaches on the Pubmed dataset \cite{pubmed}, including edge prediction \cite{hamilton2017inductive}, contrastive learning \cite{you2020graph}, and the hybrid method by combining them. We show the results in Figure \ref{fig:pilot}, and observe that contrastive learning performs the best, showing a strong transferability to the downstream task. Unfortunately, once combined with edge prediction, the performance decreases. A crucial reason is that vanilla hybrid strategies lack the knowledge to recognize the task type in the pre-training stage, which drives GNNs to be unable to distinguish tasks and further results in distortion of \emph{task-specific knowledge} due to the overlapping of different tasks.

In addition to being indistinguishable from tasks, traditional pre-training strategies on neighborhood-based GNNs cannot distinguish nodes located in different parts of
the graph but have similar neighborhoods, making models fail to learn \emph{position-specific knowledge.} and lead to suboptimal performance \cite{you2019position}. Although several recent works \cite{you2019position,sunil2021graphreach} can incorporate the position information, they require a large modification of the GNN architecture, so the flexibility is limited.

\stitle{Present work.} To address these issues, in this work, we aim to bridge the semantic gap with a unified framework for hybrid training, and enable GNNs to learn task-specific and position-specific knowledge during pre-training.
% To address the issue of hybrid training, 
Inspired by the \emph{prompt-based tuning} in natural language processing (NLP) \cite{petroni-etal-2019-language,liu2021gpt,li-liang-2021-prefix}, we propose to inject the task identification into GNNs through prompt technique in the pre-training stages. 
% Specifically, we modify the input graph by attaching a fake node (named as \textit{prompt node}) to the target node as a task token, in which the feature of the prompt node is a trainable task embedding determined by the type of each pre-training task. 
Specifically, for each given input graph, we attach a virtual node as a task token (\ie, \emph{prompt node}), in which the feature represents the trainable task embedding determined by the type of each pre-training task. 
% \zeminC{这里我们叫virtual node吧，因为fake一般是与real对应的。}
This enables GNNs to learn task-specific knowledge by conditioning on prompt nodes and thus can guarantee the coexistence of multi-angle pre-training knowledge originating from different tasks.
% keep the integrity of pre-training knowledge. 
% On the other hand, labeled instances usually differ in their 
Furthermore, to inject the position identification,
% considering most GNN architectures fail to distinguish nodes that are located in different parts of the graph and thus may lead to suboptimal performance \cite{you2019position}, 
we treat nodes at different locations as different tasks, and incorporate position information to the virtual node as a modifier of the task token. We refer to the prompt node containing two kinds of prompt information as \emph{dual prompts}, which prompt GNNs to generate task-aware and position-aware representations for the target node. We name this framework \textbf{m\underline{UL}ti-\underline{T}ask g\underline{RA}ph \underline{D}ual \underline{P}rompt}, or \model. 

Given the task-aware node representation, the learned knowledge from different pre-training tasks can be clearly distinguished, which allows for testing the transferability of each task separately. Inspired by recent research about prompt transfer \cite{vu-etal-2022-spot} in NLP, we propose a \emph{prompt-based transferability test}
% . We add prompt nodes in the fine-tuning stages, and the task embedding of the downstream task is initialized with the ones from each pre-training task to test the performance. 
% This 
to help us select the most relevant pre-training task and reduces the semantic gap significantly. Moreover, to implement the hybrid pre-training strategies, we propose a novel similarity prediction task based on $k$-nearest neighbors ($k$-NN): a node is expected to have a similar representation to a group of $k$ neighbor nodes which is sampled according to \textit{reachability}. We combine this \textit{node-group} level task with the classical edge prediction (\ie, \textit{node-node} level task), since the combination of such multi-scale tasks can help learn multi-grained richer knowledge than single-scale tasks. 

% To bridge the objective gap, we propose to unify the pre-training tasks and downstream tasks as a similarity prediction task based on our prompt framework. This is dissimilar to GPPT \cite{sun2022gppt} in which there are no explicit prompting operations and therefore cannot make the full use of the prompt's ability of integrating tasks. 
% Based on it, we propose a novel reachability prediction task and combine it with the classical edge prediction task for hybrid pre-training.

Finally, extensive experiments conducted on five widely-used datasets show that our \model, when combined with $k$-NN similarity prediction and edge prediction tasks, significantly enhances the performance of hybrid pre-training methods, and outperforms all the baseline pre-training methods. Additionally, we demonstrate the effectiveness of introducing task types and position information into prompts and show the generalizability of our proposed \model~to other pre-training tasks and GNN architectures.

% Figure \ref{fig:model} illustrates the overall framework of the proposed \model. 
The main contributions of this work are three-fold:

\begin{enumerate}[leftmargin=*] 
    \item We propose a unifying framework for graph hybrid pre-training by introducing prompt nodes in the input graph, which prompts GNNs to predict task-aware and position-aware node representation without modifying the model structure.
    \item We propose a prompt-based transferability test to find the most related pre-training task and reduce the semantic gap.
    \item We propose a novel pre-training task, a $k$-NN similarity prediction task, to increase the diversity of hybrid pre-training and learn multi-grained rich knowledge.
\end{enumerate}

\section{Related work}

\stitle{Graph Neural Networks.} Graph neural networks (GNNs) have been widely studied for their ability to extract complex semantic knowledge from graph data by utilizing recursive message passing across nodes. There are various kinds of GNN structures, such as GCN \cite{kipf2017semi}, GAT \cite{veličković2018graph}, SAGE \cite{hamilton2017inductive}, and GIN \cite{xu2018how}. Recent research also proposes transformer-based graph models without message passing mechanism \cite{dwivedi2020generalization,ying2021transformers,rampavsek2022recipe}. In this paper, our proposed model is agnostic to specific GNN structures. 

\stitle{Graph Pre-training.} The goal of graph pre-training is to capture transferable patterns across the input graph by self-supervised learning, for boosting the performance of downstream tasks \cite{panagopoulos2021transfer}. Most of the existing methods follow the \emph{pre-training and fine-tuning} scheme: pre-train GNNs with one or more pretext tasks (\eg, contrastive learning \cite{you2020graph,qiu2020gcc}, edge prediction \cite{hamilton2017inductive,sun2022gppt}, attribute masking \cite{Hu2020Strategies,jin2020self,you2020does}, and structure generation \cite{kipf2016variational,hu2020gpt}) on a pre-training dataset, which are used as an initialization for downstream tasks. The most related pre-training framework to our method is GPPT \cite{sun2022gppt}, which draws on the core idea of prompt tuning in NLP and unifies the tasks as an edge prediction problem. However, there are no explicit prompting operations on the input data in their work. As a result, it cannot take full advantage of the prompt's ability to integrate different tasks.

\stitle{Prompt Tuning.} Prompt-based tuning has become an effective way to exploit rich knowledge in a pre-trained model for various NLP tasks \cite{petroni-etal-2019-language,liu2021gpt}. With a series of tokens (namely, prompts) attached to the origin text inputs, downstream tasks are reformulated as the pre-training tasks. A prompt can be seen as a task description that prompts a pre-trained model to focus on a specific task. Recent prompts can be divided into two categories: 1) hard prompts \cite{schick-schutze-2021-exploiting,gao-etal-2021-making}, which are human-picked textual templates, and 2) soft prompts \cite{lester-etal-2021-power,li-liang-2021-prefix}, which are continuous trainable tokens. Soft prompts are more related to our work since in the graph field it's infeasible to design semantic meaningful templates. Recently, researchers started to explore the transferability of soft prompts and found that a prompt pre-trained on one or more source tasks can enhance the performance across other tasks by initialization \cite{gu-etal-2022-ppt,vu-etal-2022-spot}, which is related to our proposed prompt-based transferability test.
\section{Preliminaries}

In this section, we introduce the preliminaries of graph neural networks and the graph pre-training and fine-tuning scheme.

\subsection{Graph Neural Networks}

Let triple $G=(\mathcal{V}, \mathcal{E}, \mX)$ denote an undirected graph, where $\mathcal{V}$ is the set of nodes, $\mathcal{E}\in\mathbb R^{|\mathcal{V}| \times |\mathcal{V}|}$ is the set of edges, and $\mX\in\mathbb R^{|\mathcal{V}|\times d}$ is the node feature matrix where $d$ is the feature dimension. Graph neural networks (GNNs) \cite{bruna2013spectral,gilmer2017neural} aim to calculate the representation of the target node with the local neighborhood information. Formally, let $\phi(\cdot;\theta)$ denote an $L$-layer GNN architecture parameterized by $\theta$. In the $l$-th layer, the representation of node $v$, \emph{i.e.}, $\vh^l_v\in\mathbb{R}^{d_l}$, can be calculated by
\begin{equation*}
    \vh_v^l = \textsc{Aggr}\left(\vh_v^{l-1},\{\vh_i^{l-1} \mid i \in \mathcal{N}_v\}; \theta^l\right),\quad l=1,2,\cdots,L,
\end{equation*}
where $\mathcal{N}_v$ is the neighbor set of node $v$, and $\textsc{Aggr}(\cdot;\theta^l)$ is the neighborhood aggregation function parameterized by $\theta^l$ in layer $l$. The initial representation $\vh^0_v$ is defined as the node feature $\vx_v \in\mathbb R^{d}$. 

Additionally, different GNN architectures may differ in the neighborhood aggregation function $\textsc{Aggr}$ \cite{veličković2018graph,xu2018how}. In particular, our proposed \model~takes effect in the input graph and is agnostic to the model structures, and we aim to make it flexible to most neighborhood aggregation-based GNNs.

\subsection{Pre-train and Fine-tune GNNs}

Supervised node representation learning in large-scale graphs typically requires a significant amount of annotated data for each individual downstream task. However, the process of manually labeling a large number of nodes can be resource-intensive and time-consuming. Besides, training GNNs from scratch for each downstream task can also be inefficient. To address these challenges, the typical framework of \emph{pre-training} and \emph{fine-tuning} has been proposed as a potential solution \cite{hu2020gpt,lu2021learning,you2020graph}. This approach utilizes easily accessible information to encode the intrinsic semantics of the graph, and the derived pre-trained GNNs can then serve as initialization for generalization to other downstream tasks.

Formally, a graph pre-training process can be written as follows:
\begin{equation*}
    \theta^{\text{pre}} =\arg\min\limits_{ \theta} \mathcal{L}_{\text{pre}} (G; 
    \theta),
\end{equation*}
where $\mathcal{L}_{\text{pre}}$ is the objective of interest. To catch semantic properties from different views and enrich the knowledge, a prevalent method is hybrid pre-training, which integrates various training objectives in a multi-task learning fashion, formulated as:
\begin{equation*}
    \theta^{\text{pre}} = \arg\min\limits_{ \theta} \sum_{i=1}^N w_i \mathcal{L}^i_{\text{pre}} (G; 
    \theta),
\end{equation*}
where $\mathcal{L}^i_{\text{pre}}$ is the $i$-th pre-training objective, $w_i$ is the corresponding weight and $N$ is the number of objectives.

Node classification \cite{jin2020self} is a widely-used downstream task. When fine-tuning pre-trained GNNs to this task, a new projection head $f(\cdot; \Phi)$ parameterized by $\Phi$ is usually used to map the node representation to labels. The process can be formulated as:

\begin{equation*}
    \theta^*, \Phi^* = \arg\min\limits_{ \theta, \Phi} \sum_{v} \mathcal{L}_{\text{down}} (f(\vh(v; \theta); \Phi), l_v),
\end{equation*}
where $l_v$ is the ground-truth label of node $v$ and $\mathcal{L}_{\text{down}}$ is the downstream loss function, \eg, cross-entropy for node classification. $\vh(v; \theta)$ is the node representation from a GNN parameterized by $\theta$, which is initialized by $\theta^{\text{pre}}$.

\newcommand{\distance}[2]{D\left(#1, #2\right)}
\newcommand{\similarity}[2]{S\left(#1, #2\right)}

\section{Multi-task Graph dual Prompt}

In this section, we introduce our framework for hybrid pre-training. We first discuss how to leverage the prompt technique in NLP to incorporate information missing from the current hybrid pre-training methods (\sref{sec:prompt}), then introduce each component inside the prompts (\sref{sec:prompt_task}). Next, we propose a transferability test based on prompts to find the most relevant pre-training task, to reduce the semantic gap (\sref{sec:prompt_transfer}). Finally, we summarize the overall pipeline of our proposed framework (\sref{sec:prompt_summary}).

\subsection{Prompt Nodes}

\label{sec:prompt}

Although the hybrid pre-training with neighborhood-based GNNs can learn multi-grained knowledge which is beneficial to reduce the semantic gap from downstream tasks, there are still two major challenges: 1) hybrid pre-training cannot distinguish task types, leading to the knowledge learned from different tasks being mixed together and interfering with each other; and 2) neighborhood-based GNNs cannot distinguish nodes located in different parts of the graph, making GNNs fail to learn position-specific knowledge. Recent efforts require a large modification of the GNNs to incorporate position information \cite{you2019position,sunil2021graphreach}, which limits the flexibility of the model structure.

\begin{figure}[t]
    \centering
    \includegraphics[width=0.47\textwidth]{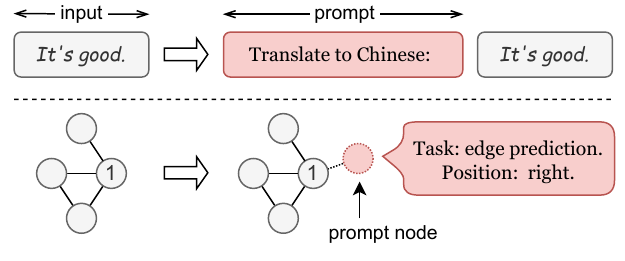}
    \caption{Prompt tuning examples in the NLP field (up) and in our work (down).}
    \label{fig:prompt}
\end{figure}

To address these issues, inspired by prompt tuning \cite{petroni-etal-2019-language,liu2021gpt,schick-schutze-2021-exploiting}, we propose to inject the indicative information through prompts. We illustrate it in Figure \ref{fig:prompt}. In NLP, a prompt is typically a task-specific textual template prepended to the input sentence, which prompts the pre-trained model to execute the specific task, \eg, a translation task. Motivated by this, we propose to add a virtual node (\ie, \textbf{prompt node}) describing task and position information, with a link to the target node that we want to obtain the representation of. For example, in Figure~\ref{fig:prompt}, when calculating the representation for node 1, we attach a prompt node encoding the task (\textit{edge prediction}) and the position (in the \textit{right} part of the graph) to it and use the modified graph as the input. Note that this transformation takes effect in the input space without changing the model framework. 

We name this pre-training framework \textbf{m\underline{UL}ti-\underline{T}ask g\underline{RA}ph \underline{D}ual \underline{P}rompt} (\model). The features of prompt node, \ie, $\vp \in \mathbb R^d$, consists of dual parts, task embedding and position embedding. We describe their details in the following.

\subsection{Prompt Feature}

\label{sec:prompt_task}
\subsubsection{Task Embedding}

The task embedding is a series of trainable embeddings $\theta^{\text{task}}=\{\vp^{\text{task} (1)}, \vp^{\text{task} (2)}, \cdots, \vp^{\text{task} (N)}\}$ for $N$ pre-training tasks, respectively. In the pre-training stage, we randomly select a task $i\in[N]$ and incorporate the corresponding task embedding $\vp^{\text{task} (i)}$ to prompt nodes, which is trained together with the base GNN models. After pre-training, task embeddings can store the task-specific semantic knowledge and the GNN stores the task-independent global knowledge, which keeps the integrity of pre-training. 

\subsubsection{Position Embedding} 

\label{sec:prompt_pos}

The position of a node in the graph can be expressed as a series of distances to some fixed reference nodes, or called anchors \cite{you2019position,sunil2021graphreach}. Previous studies have shown the effectiveness of using \textit{reachability} as a form of distance, which is holistic, semantic, and robust \cite{sunil2021graphreach}. A $t-$reachability $\reach{i}{j}{t}$ from node $v_i$ to node $v_j$ is defined as the probability of starting from $v_i$ and reaching $v_j$ through a $t$-step random walk. We provide further calculation details about reachability and the complexity analysis in Appendix \ref{sec:appendix_reachability}.

To obtain high-quality position encodings, anchors should be selected carefully. Instead of selecting them randomly \cite{you2019position}, we calculate the total reachability for each node by summing the reachability from all the other nodes in the graph to itself, and select the top-$m$ nodes with the largest total reachability as the anchors, \ie,
\begin{equation}
    \label{eq:anchor}
    \va = \underset{v_i\in \mathcal{V}}{\arg \text{sort}} \sum_{v_j\in \mathcal{V}} \reach{j}{i}{t},
\end{equation}
where $t$ and $m$ are hyper-parameters. The anchors are the most \emph{representative} and \textit{reachable} nodes in $t$ steps in the graph. Based on it, the position encoding of a node $v_i$ can be formulated as:
\begin{equation}
    \label{eq:position_encoding}
    \widetilde \vp^{\text{pos}}_i = \left(
        \reach{i}{a_1}{t}, \reach{i}{a_2}{t}, \cdots,
        \reach{i}{a_m}{t}
    \right),
\end{equation}
where $a_j$ is the $j$-th element in the anchors $\va$. 

The position encoding describes the reachabilities from the node to the top-$m$ anchors which determines the node position. Since the elements in position encoding might be small when the random walk step $t$ is large, and the dimensionality $t$ does not match the prompt feature, we normalize the position encoding by dividing the standard deviation and project it into $\mathbb R^d$ space with a linear layer and $\text{tanh}$ activation, to obtain the final position embedding:
\begin{equation}
    \label{eq:position_embedding}
    \vp^{\text{pos}}_i = \text{tanh}\left(
        \mW^{\text{pos}} \frac{\widetilde \vp^{\text{pos}}_i}{\sigma_i + \epsilon} + \vb^{\text{pos}}
    \right),
\end{equation}
where $\sigma_i$ is the standard deviation of $\widetilde \vp^{\text{pos}}_i$ and $\epsilon$ is a small number to keep numerically stable. $\theta^{\text{pos}} = \{\mW^{\text{pos}}\in\mathbb R^{d\times m},  \vb^{\text{pos}}\in\mathbb R^d\}$ are weights, which are learned by pre-training and fine-tuned for the downstream task.

\subsubsection{Combining Task and Position Embedding}

\label{sec:prompt_combine}

The prompt feature is the weighted sum of the two embeddings. For a node $v_i$ with the $j$-th pre-training task, the prompt feature is:
\begin{equation}
    \label{eq:prompt}
    \vp_i = \vp^{\text{task}(j)} + w^{\text{pos}}\vp^{\text{pos}}_i,
\end{equation}
where $w^{\text{pos}}$ is a hyper-parameter controlling the strength of position embedding. 

In the NLP field, prompts and normal tokens can be naturally concatenated together as input since they are in the same word space. However, for graph data, the node features can be discrete or sparse which is different from the continuous prompt feature. Simply connecting two nodes with features in different spaces directly might lead to unstable training. To better align the prompt feature $\vp_i$ and normal node feature $\vx_i$, we transform them using distinct linear layers first:
\begin{equation*}
    \vh_i = \begin{cases}
        \text{Linear}(\vp_i; \theta^{\text{prompt}}) & \mbox{if $v_i$ is a prompt node,}\\
        \text{Linear}(\vx_i; \theta^{\text{normal}}) & \mbox{if $v_i$ is a normal node,}\\
    \end{cases}
\end{equation*}
where $\vh_i$ is the representation for node $v_i$ to feed the GNN. $\theta^{\text{prompt}}$ and $\theta^{\text{normal}}$ are parameters for transforming the prompt feature and normal features into the same feature space, respectively.

\setlength{\textfloatsep}{3pt}

\begin{algorithm}[t]
    \caption{\textsc{Prompting}}
    \label{algo:prompting}
    \LinesNumbered
    \KwIn{Input graph $G=(\mathcal{V}, \mathcal{E}, \mX)$, Target node sets $\{v_1, \cdots, v_n\}$, Current task embedding $p^{\text{task}(j)}$}
    \KwOut{Prompted graph $G'$}
    $\mathcal{V}'\leftarrow \mathcal{V}, \mathcal{E}' \leftarrow{\mathcal{E}}, \mX'\leftarrow{\mX}$\;
    \For{$i=1$ \KwTo $n$}{
        Create a prompt node $v_i^p$ for the target node $v_i$\;
        Calculate prompt feature $p_i$ by \Eqref{eq:anchor}-\Eqref{eq:prompt}\;
        $\mathcal{V}'\leftarrow \mathcal{V}'\cup \left\{v_i^p\right\}$\;
        $\mathcal{E}'\leftarrow \mathcal{E}' \cup \left\{\left(v_i^p, v_i\right)\right\}$\;
        $\mX'\leftarrow \mX' \cup \left\{p_i\right\}$\;
    }
    \Return{$G'=(\mathcal{V}', \mathcal{E}', \mX')$}
\end{algorithm}

\begin{algorithm}[t]
    \caption{\textsc{Pre-training Framework of \model}}
    \label{algo:pre-training}
    \LinesNumbered
    \KwIn{Input graph $G=(\mathcal{V}, \mathcal{E}, \mX)$, Pre-training tasks $\{t_1, \cdots, t_N\}$, Sub-graph sampler $\textsc{Sub-graph}(\cdot)$}
    \KwOut{Pre-trained parameters $\theta$}
    Initialize GNN parameters $\theta^{\text{GNN}}$, task embeddings $\theta^{\text{task}}$, position embedding parameters $\theta^{\text{pos}}$, feature transformer parameters $\theta^{\text{prompt}}$ and $\theta^{\text{normal}}$\;
    \Repeat{model convergences}{
        Sample a task $t_j\in \{t_1, \cdots, t_N\}$\;
        Sample target nodes $\{v_1, \cdots, v_n\}\subset\mathcal{V}$ required by $t_j$\;
        $G'\leftarrow \textsc{Sub-graph}(\{v_1, \cdots, v_n\}, G)$\;
        $G'\leftarrow \textsc{Prompting}\left(
            G', \{v_1, \cdots, v_n\}, p^{\text{task}(j)}
        \right)$\;
        Optimize parameters on $G'$ for the task $t_j$\;
    }
    \Return{$\theta=\left(
        \theta^{\text{GNN}}, \theta^{\text{task}}, \theta^{\text{pos}}, \theta^{\text{prompt}}, \theta^{\text{normal}}
    \right)$}
\end{algorithm}

\begin{figure*}[htbp]
    \centering
    \includegraphics[width=0.75\textwidth]{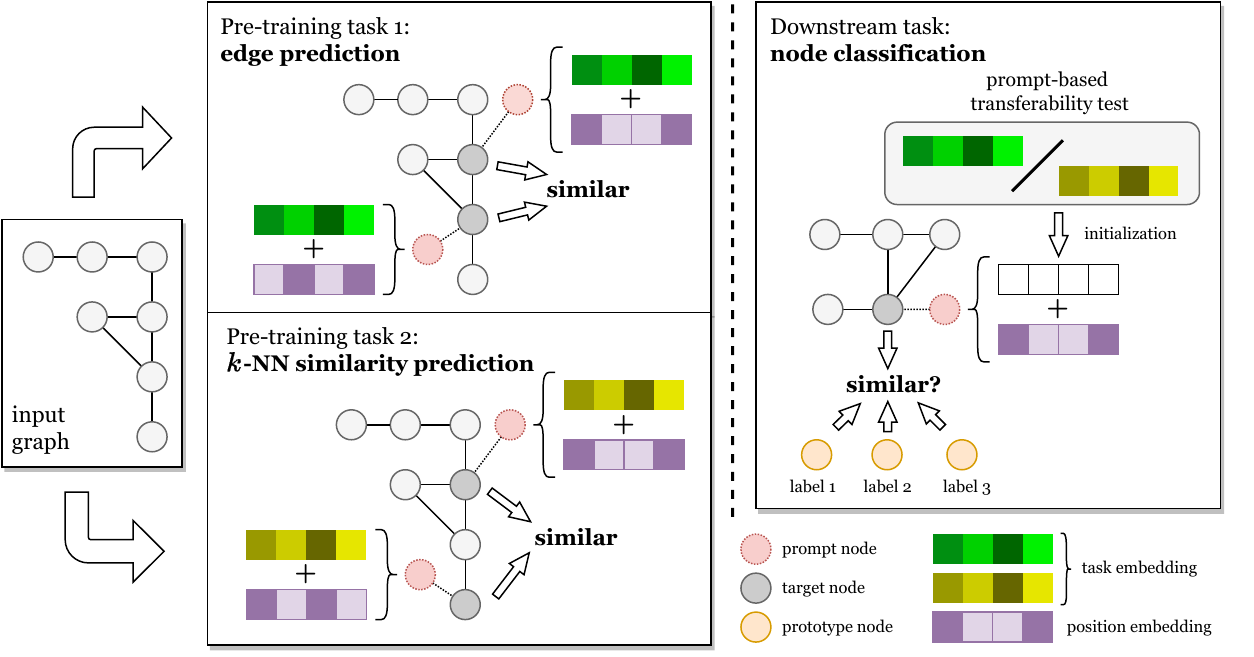}
    \caption{An illustration of the proposed \model. We use dual prompt nodes to incorporate task and position information. The prompt-based transferability test is proposed to reduce the semantic gap. 
    % Moreover, we unify downstream tasks and upstream tasks as similarity prediction problems to bridge the objective gap.
    }
    \label{fig:model}
\end{figure*}

\subsection{Prompt-based Transferability Test}

\label{sec:prompt_transfer}

For the downstream task, position embedding can be obtained by the same method as the pre-training stage, while how to obtain task embedding is a question since downstream and upstream tasks are usually different. Training task embeddings for each new downstream task from scratch is not only inefficient, but also cannot fully utilize the knowledge contained in the pre-trained task embeddings.

Motivated by the recent progress of the prompt transfer and prompt pre-training \cite{vu-etal-2022-spot} in NLP, we argue that the knowledge can be transferred through the initialization of task embeddings. Particularly, we propose a \textbf{prompt-based transferability test}: using each pre-trained task embedding to initialize the one for the downstream task, and fine-tuning the whole model in the training set. Finally, we test the performance of each initialization in the validation set and adopt the best model. In other words, we regard the choice of initialization strategies from each pre-training task as a hyper-parameter. This helps us select the most relevant pre-training task and reduces the semantic gap significantly. 

Compared to pre-training multiple models with different pre-training strategies and finding the most relevant one, our \model~is less time-consuming and space-consuming since we only need to pre-train once and obtain a single GNN weight. Moreover, our hybrid framework can take full advantage of each task and gain diverse knowledge from multiple views. We find empirically that our \model~outperforms every single pre-training method consistently, regardless of the initialization of task embedding (\sref{sec:experiment_results}).

\subsection{Overall Pipeline of \model}

\label{sec:prompt_summary}

In this section, we summarize the overall pipeline and algorithms of our proposed \model~framework. 

\stitle{Prompting.} Algorithm \ref{algo:prompting} illustrates the prompting function. Given a graph $G$, a set of target nodes to prompt and current task embedding, we first initialize prompted graph as $G$, corresponding to line 1. For each target node $v_i$, we create a prompt node and calculate its feature by \Eqref{eq:anchor}-\Eqref{eq:prompt} with the current task embedding in line 3-4. In line 5-7, we add the prompt node to the prompted graph, with an edge between it and the target node.

\stitle{Pre-training. } The overall pre-training framework of \model~is illustrated in Algorithm \ref{algo:pre-training}. In line 1, we initialize all parameters. Next, we pre-train \model~using multiple tasks. We first sample a task in line 3, and sample target nodes required by this task in line 4. In line 5, we sample a sub-graph from these target nodes, to prevent graphs from being too large to fit into the device. Particularly, we follow \cite{hu2020gpt} and sample a dense sub-graph by using the \method{LADIES} algorithm \cite{zou2019layer}, which ensures that the sampled nodes are highly interconnected with each other and keep the structural information. Next, we prompt the sub-graph using Algorithm \ref{algo:prompting} and optimize all parameters in line 6-7.

\stitle{Fine-tuning. } We further present the algorithm for fine-tuning in Appendix \sref{sec:appendix_ft}.

\section{Model Implementation}

So far, we have shown that our proposed \model~framework is able to boost the hybrid pre-training methods by introducing prompts. Furthermore, the prompt technique is usually used for bridging the training objective gap between upstream and downstream tasks in NLP. Based on it, we unify the downstream and upstream task as a \textit{similarity prediction} task in our \model~framework, to bridge the objective gap. Similarly, Sun \etal~\cite{sun2022gppt} also proposed to unify the downstream and upstream tasks as an edge prediction problem. Here we extend it to similarity prediction to incorporate multiple pre-training tasks. The similarity in our work is defined as the cosine similarity function: 
\begin{equation}
    \label{eq:similarity}
    S(\vh_i, \vh_j) = \frac{\vh_i}{||\vh_i||_2}\cdot \frac{\vh_j}{||\vh_j||_2}.
\end{equation}

The first problem is how to design multiple tasks that may benefit downstream tasks. We propose to incorporate tasks at multiple scales to learn multi-grained rich knowledge, including the similarity prediction between nodes (\ie, node-node level), and the similarity between one node and a group of nodes (\ie, node-group level). Empirically, we find that such heterogeneous pre-training tasks can outperform single-scale ones (\sref{sec:experiment_results}).

Next, we introduce two pre-training tasks used for \model: edge prediction and $k$-NN similarity prediction. Figure \ref{fig:model} illustrates the overall model architecture.

\subsection{Edge Prediction}

\label{sec:edge}

To implement the node-node level task, we adopt a typical pre-training task: edge prediction. In this task, we randomly select a node $v$ which has at least one neighbor, and sample a node $v^+$ from its neighbors, as well as another node $v^-$ without an edge between $v^-$ and $v$. The task is to maximize the similarity between $v$ and $v^+$ and minimize the similarity between $v$ and $v^-$. Suppose the node representations of $v$, $v^+$ and $v^-$ given by the GNN are $\vh$, $\vh^+$ and $\vh^-$ respectively. The objective loss is formulated as follows:
\begin{equation*}
    \mathcal{L}^{\text{edge}} = -S(\vh, \vh^+) + \max\{0, S(\vh, \vh^-) - \alpha \},
\end{equation*}
where $\alpha$ is a hyper-parameter controlling the margin. A larger $\alpha$ leads to a lower similarity between $\vh$ and $\vh^-$.

\subsection{$k$-Nearest Neighbors Similarity Prediction}

\label{sec:reachability}

For the node-group level task, we design a novel pre-training strategy based on $k$-nearest neighbors ($k$-NN) strategy: maximize the similarity between a node and its $k$ nearest neighbors (positive samples), and minimize the similarity between it and other $k$ further nodes (negative samples). As mentioned in \sref{sec:prompt_pos}, reachability is a good metric of distance for neighbor selection, therefore, we propose to push a node with its $k$ neighbors with the largest reachability together since they share more similar semantics. A similar idea is proposed by Xu \etal~\cite{Xu2020Graph} in which reachability is one of the components to compute node similarity scores. Compared to edge prediction, $k$-NN focuses on \emph{multi-hop} connection in the graph, which is a generalization of edge prediction that focuses on \emph{one-hop} connection between nodes. Combining them together could express more graph structure semantics comprehensively.

Furthermore, instead of using fixed positive examples, we propose to sample different positive examples for a node using the reachabilities as probabilities in different batches, which helps to broaden the contrastive scope. Formally, for a anchor node $v_i$, we sample $k$ positive examples by:
\begin{equation*}
    x^{1+}_i, x^{2+}_i, \cdots, x^{k+}_i \sim \text{Cat}\left(
        \reach{i}{1}{t'}, \cdots, \reach{i}{|\mathcal{V}|}{t'}
    \right),
\end{equation*}
where $\text{Cat}$ is categorical distribution and $t'$ is a hyper-parameter about the step of random walks. Similarly, the negative examples are sampled with the reciprocal of reachabilities, formulated as:
\begin{equation*}
    x^{1-}_i, x^{2-}_i, \cdots, x^{k-}_i \sim \text{Cat}\left(
        \frac{Z_i}{\reach{i}{1}{t'}}, \cdots, \frac{Z_i}{\reach{i}{|\mathcal{V}|}{t'}}
    \right),
\end{equation*}
where $Z_i$ is used to normalize values to valid probabilities. Suppose the node representation for anchor node $v_i$ is $\vh_i$, the ones for positive examples are $P=\{\vh_i^{1+}, \cdots, \vh_i^{k+}\}$, and the ones for negative examples are $N=\{\vh_i^{1-}, \cdots, \vh_i^{k-}\}$. We utilize triplet loss \cite{schroff2015facenet} to encourage the GNN to find embeddings where the minimum distance between $\vh_i$ and negative examples is larger than the maximum distance between $\vh_i$ and positive examples plus the margin parameter $\alpha'$, formally,
\begin{align*}
    \mathcal{L}_{\text{triplet}}^{\text{reach}}& = \max\{
        0, J
    \}^2,\\
    J&=\max_{j=1}^k\left\{
        \distance{\vh_i}{\vh_i^{j+}}
    \right\} + \max_{j=1}^k\left\{
        \alpha' - \distance{\vh_i}{\vh_i^{j-}}
    \right\},
\end{align*}
where $D(\vh_1, \vh_2) = 1 - S(\vh_1, \vh_2)$ is the distance function based on similarity function in \Eqref{eq:similarity}. Given that the $\max$ function is non-smooth which causes the network to converge to a bad local optimum \cite{oh2016deep}, we optimize the following smooth upper bound:
\begin{align*}
    {\mathcal{\widetilde L}}_{\text{triplet}}^{\text{reach}}& = \max\{
        0, \widetilde J
    \}^2,\\
    \widetilde J&=\log\sum_{j=1}^k
        \exp\left(
            \distance{\vh_i}{\vh_i^{j+}}
        \right)
     + \log\sum_{j=1}^k
        \exp\left(
            \alpha' - \distance{\vh_i}{\vh_i^{j-}}
        \right),
\end{align*}

Besides, to penalize the distances between positive examples and encourage the GNN to generate similar embeddings for positive examples, we further incorporate center loss \cite{wen2016discriminative}, formulated as:
\begin{align*}
    \mathcal{L}_{\text{center}}^{\text{reach}} = \frac{1}{k}\sum_{j=1}^k D^2\left({\vh_i}, \vh_i^{j+}\right).
\end{align*}
Finally, we sum these two objectives: ${\mathcal{\widetilde L}}_{\text{triplet}}^{\text{reach}}+ \mathcal{L}_{\text{center}}^{\text{reach}}$, as the objective loss for the $k$-NN similarity prediction.

\subsection{Downsteam Node Classification}

\label{sec:downstream}

Similar to \cite{sun2022gppt}, we formulate the downstream node classification as a similarity prediction problem. For the total $C$ classes, we use $C$ prototype nodes for each class, defined by $E=[e_1, \cdots, e_C]^\top \in \mathbb R^{C\times d'}$, where $d'$ is the dimension of node representation of the GNN. The classification is performed by querying the similarity between node representation and prototype nodes. Based on InforNCE loss \cite{oord2018representation} which is widely used in contrastive learning, we adopt the following loss to fine-tune GNNs:
\begin{align*}
    \mathcal{L}_{\text{down}}& = -\sum_{v} \log \frac{
        \exp \similarity{\vh}{e_c}
    }{
        \sum_{i=1}^C \exp \similarity{\vh}{e_i}
    },
\end{align*}
where $\vh$ and $c$ are the node representation and class label for node $v$, respectively.

\section{Experiments}

% table* generated by Excel2LaTeX from sheet 'graph'
\begin{table*}[htbp]
  \centering
  \caption{Micro-F1 (\%) results on the five datasets. Bold: best results. \underline{Underline}: best results among \method{Edge}-based and \method{$k$-NN}-based methods, \emph{i.e.}, \method{Edge}, \method{$k$-NN}, \method{Hybrid} and \model. }
% Table generated by Excel2LaTeX from sheet 'graph'
\begin{tabular}{lccccccccccc}
\toprule[1pt]
\multirow{2}[4]{*}{} & \multicolumn{3}{c}{\textbf{DBLP}} &      & \multicolumn{3}{c}{\textbf{Pubmed}} &      & \multicolumn{3}{c}{\textbf{CoraFull}} \\
\cmidrule{2-4}\cmidrule{6-8}\cmidrule{10-12}     & 8-shot & 32-shot & 128-shot &      & 8-shot & 32-shot & 128-shot &      & 8-shot & 32-shot & 128-shot \\
\midrule[1pt]
\method{No Pre-train} & 65.0\textsubscript{$\pm$0.4} & 68.0\textsubscript{$\pm$0.4} & 73.3\textsubscript{$\pm$0.7} &      & 69.6\textsubscript{$\pm$0.8} & 76.0\textsubscript{$\pm$0.1} & 79.8\textsubscript{$\pm$0.5} &      & 38.3\textsubscript{$\pm$0.3} & 48.2\textsubscript{$\pm$0.7} & 50.9\textsubscript{$\pm$0.4} \\
\method{CL} & \boldmath{}\textbf{71.7\textsubscript{$\pm$0.6}}\unboldmath{} & 72.2\textsubscript{$\pm$0.9} & 77.3\textsubscript{$\pm$0.6} &      & 73.5\textsubscript{$\pm$2.7} & 79.3\textsubscript{$\pm$0.9} & 82.0\textsubscript{$\pm$0.3} &      & 34.6\textsubscript{$\pm$0.6} & 45.3\textsubscript{$\pm$0.3} & 46.7\textsubscript{$\pm$0.4} \\
\method{GPT-GNN} & 70.4\textsubscript{$\pm$0.9} & 73.1\textsubscript{$\pm$0.8} & 77.9\textsubscript{$\pm$0.2} &      & 73.0\textsubscript{$\pm$1.7} & 79.8\textsubscript{$\pm$0.3} & 82.4\textsubscript{$\pm$0.2} &      & 32.4\textsubscript{$\pm$1.4} & 42.8\textsubscript{$\pm$1.3} & 44.7\textsubscript{$\pm$2.1} \\
\method{GPPT} (\method{Edge}) & 68.8\textsubscript{$\pm$1.4} & 71.3\textsubscript{$\pm$0.9} & 76.6\textsubscript{$\pm$0.1} &      & 73.2\textsubscript{$\pm$0.9} & 78.5\textsubscript{$\pm$0.4} & 80.2\textsubscript{$\pm$0.1} &      & 37.6\textsubscript{$\pm$0.4} & 46.8\textsubscript{$\pm$0.7} & 49.0\textsubscript{$\pm$0.3} \\
\method{$k$-NN} & 68.6\textsubscript{$\pm$0.5} & 69.5\textsubscript{$\pm$0.8} & 75.4\textsubscript{$\pm$0.3} &      & 75.0\textsubscript{$\pm$2.2} & 80.3\textsubscript{$\pm$0.3} & 81.4\textsubscript{$\pm$0.2} &      & 36.2\textsubscript{$\pm$0.7} & 46.8\textsubscript{$\pm$0.2} & 49.1\textsubscript{$\pm$0.8} \\
\method{Hybrid} (\method{Edge} + \method{$k$-NN}) & 68.9\textsubscript{$\pm$0.5} & 70.4\textsubscript{$\pm$1.1} & 76.3\textsubscript{$\pm$0.7} &      & 73.7\textsubscript{$\pm$0.5} & 79.2\textsubscript{$\pm$0.5} & 81.3\textsubscript{$\pm$0.3} &      & 37.5\textsubscript{$\pm$0.6} & 48.1\textsubscript{$\pm$1.1} & 50.5\textsubscript{$\pm$0.5} \\
\midrule
\model~(\method{Edge} + \method{$k$-NN}) & \underline{71.6\textsubscript{$\pm$1.3}} & \underline{\boldmath{}\textbf{73.6\textsubscript{$\pm$0.7}}\unboldmath{}} & \underline{\boldmath{}\textbf{78.5\textsubscript{$\pm$0.7}}\unboldmath{}} &      & \underline{\boldmath{}\textbf{75.6\textsubscript{$\pm$1.4}}\unboldmath{}} & \underline{\boldmath{}\textbf{80.5\textsubscript{$\pm$0.6}}\unboldmath{}} & \underline{\boldmath{}\textbf{83.1\textsubscript{$\pm$0.3}}\unboldmath{}} &      & \underline{\boldmath{}\textbf{41.8\textsubscript{$\pm$1.6}}\unboldmath{}} & \underline{\boldmath{}\textbf{52.1\textsubscript{$\pm$1.7}}\unboldmath{}} & \underline{\boldmath{}\textbf{54.8\textsubscript{$\pm$0.8}}\unboldmath{}} \\
 \quad- init with \method{Edge} & \textcolor[rgb]{ .682,  .667,  .667}{70.2\textsubscript{$\pm$1.0}} & \textcolor[rgb]{ .682,  .667,  .667}{73.6\textsubscript{$\pm$0.7}} & \textcolor[rgb]{ .682,  .667,  .667}{78.5\textsubscript{$\pm$0.7}} & \textcolor[rgb]{ .682,  .667,  .667}{} & \textcolor[rgb]{ .682,  .667,  .667}{74.4\textsubscript{$\pm$2.2}} & \textcolor[rgb]{ .682,  .667,  .667}{81.0\textsubscript{$\pm$0.8}} & \textcolor[rgb]{ .682,  .667,  .667}{82.7\textsubscript{$\pm$0.4}} & \textcolor[rgb]{ .682,  .667,  .667}{} & \textcolor[rgb]{ .682,  .667,  .667}{41.8\textsubscript{$\pm$1.6}} & \textcolor[rgb]{ .682,  .667,  .667}{52.0\textsubscript{$\pm$1.7}} & \textcolor[rgb]{ .682,  .667,  .667}{54.8\textsubscript{$\pm$0.8}} \\
 \quad- init with \method{$k$-NN} & \textcolor[rgb]{ .682,  .667,  .667}{71.6\textsubscript{$\pm$1.3}} & \textcolor[rgb]{ .682,  .667,  .667}{74.1\textsubscript{$\pm$1.0}} & \textcolor[rgb]{ .682,  .667,  .667}{78.2\textsubscript{$\pm$0.5}} & \textcolor[rgb]{ .682,  .667,  .667}{} & \textcolor[rgb]{ .682,  .667,  .667}{75.6\textsubscript{$\pm$1.4}} & \textcolor[rgb]{ .682,  .667,  .667}{80.5\textsubscript{$\pm$0.6}} & \textcolor[rgb]{ .682,  .667,  .667}{83.1\textsubscript{$\pm$0.3}} & \textcolor[rgb]{ .682,  .667,  .667}{} & \textcolor[rgb]{ .682,  .667,  .667}{42.4\textsubscript{$\pm$3.2}} & \textcolor[rgb]{ .682,  .667,  .667}{52.1\textsubscript{$\pm$1.7}} & \textcolor[rgb]{ .682,  .667,  .667}{53.0\textsubscript{$\pm$2.8}} \\
\midrule
\midrule
\multirow{2}[4]{*}{} & \multicolumn{3}{c}{\textbf{Coauthor-CS}} &      & \multicolumn{3}{c}{\textbf{ogbn-arxiv}} &      & \multicolumn{3}{c}{\textbf{\emph{Average}}} \\
\cmidrule{2-4}\cmidrule{6-8}\cmidrule{10-12}     & 8-shot & 32-shot & 128-shot &      & 8-shot & 32-shot & 128-shot &      & 8-shot & 32-shot & 128-shot \\
\midrule[1pt]
\method{No Pre-train} & 80.4\textsubscript{$\pm$0.3} & 84.5\textsubscript{$\pm$0.1} & 86.5\textsubscript{$\pm$0.1} &      & 28.2\textsubscript{$\pm$0.7} & 32.7\textsubscript{$\pm$1.0} & 43.2\textsubscript{$\pm$1.7} &      & 56.3 & 61.9 & 66.7 \\
\method{CL} & 84.0\textsubscript{$\pm$0.7} & 87.1\textsubscript{$\pm$0.4} & 88.6\textsubscript{$\pm$0.2} &      & 32.4\textsubscript{$\pm$0.9} & \boldmath{}\textbf{39.1\textsubscript{$\pm$0.8}}\unboldmath{} & 45.5\textsubscript{$\pm$1.8} &      & 59.3 & 64.6 & 68.0 \\
\method{GPT-GNN} & 82.5\textsubscript{$\pm$0.5} & 86.9\textsubscript{$\pm$0.6} & 88.8\textsubscript{$\pm$0.5} &      & \boldmath{}\textbf{34.0\textsubscript{$\pm$1.0}}\unboldmath{} & 33.9\textsubscript{$\pm$1.1} & 37.0\textsubscript{$\pm$1.1} &      & 58.5 & 63.3 & 66.1 \\
\method{GPPT} (\method{Edge}) & 81.0\textsubscript{$\pm$0.6} & 85.6\textsubscript{$\pm$0.5} & 88.3\textsubscript{$\pm$0.2} &      & \underline{28.7\textsubscript{$\pm$0.8}} & 32.2\textsubscript{$\pm$1.1} & 42.6\textsubscript{$\pm$0.2} &      & 57.8 & 62.9 & 67.3 \\
\method{$k$-NN} & 80.2\textsubscript{$\pm$0.5} & 85.6\textsubscript{$\pm$0.2} & 88.5\textsubscript{$\pm$0.2} &      & 25.9\textsubscript{$\pm$0.5} & 33.3\textsubscript{$\pm$1.3} & 42.5\textsubscript{$\pm$0.3} &      & 57.2 & 63.1 & 67.4 \\
\method{Hybrid} (\method{Edge} + \method{$k$-NN}) & 81.3\textsubscript{$\pm$0.4} & 85.6\textsubscript{$\pm$0.1} & 88.1\textsubscript{$\pm$0.2} &      & 27.8\textsubscript{$\pm$4.3} & 28.7\textsubscript{$\pm$1.0} & 39.0\textsubscript{$\pm$0.3} &      & 57.8 & 62.4 & 67.0 \\
\midrule
\model~(\method{Edge} + \method{$k$-NN}) & \underline{\boldmath{}\textbf{84.7\textsubscript{$\pm$0.3}}\unboldmath{}} & \underline{\boldmath{}\textbf{89.5\textsubscript{$\pm$0.1}}\unboldmath{}} & \underline{\boldmath{}\textbf{91.4\textsubscript{$\pm$0.4}}\unboldmath{}} &      & 28.3\textsubscript{$\pm$0.2} & \underline{35.7\textsubscript{$\pm$0.5}} & \underline{\boldmath{}\textbf{47.4\textsubscript{$\pm$1.1}}\unboldmath{}} &      & \underline{\textbf{60.4}} & \underline{\textbf{66.3}} & \underline{\textbf{71.0}} \\
 \quad- init with \method{Edge} & \textcolor[rgb]{ .682,  .667,  .667}{84.7\textsubscript{$\pm$0.3}} & \textcolor[rgb]{ .682,  .667,  .667}{89.5\textsubscript{$\pm$0.1}} & \textcolor[rgb]{ .682,  .667,  .667}{91.4\textsubscript{$\pm$0.4}} & \textcolor[rgb]{ .682,  .667,  .667}{} & \textcolor[rgb]{ .682,  .667,  .667}{28.3\textsubscript{$\pm$0.2}} & \textcolor[rgb]{ .682,  .667,  .667}{36.9\textsubscript{$\pm$0.6}} & \textcolor[rgb]{ .682,  .667,  .667}{46.8\textsubscript{$\pm$0.9}} &      & \textcolor[rgb]{ .682,  .667,  .667}{59.9} & \textcolor[rgb]{ .682,  .667,  .667}{66.6} & \textcolor[rgb]{ .682,  .667,  .667}{70.8} \\
 \quad- init with \method{$k$-NN} & \textcolor[rgb]{ .682,  .667,  .667}{84.1\textsubscript{$\pm$0.7}} & \textcolor[rgb]{ .682,  .667,  .667}{89.1\textsubscript{$\pm$0.5}} & \textcolor[rgb]{ .682,  .667,  .667}{90.9\textsubscript{$\pm$0.4}} & \textcolor[rgb]{ .682,  .667,  .667}{} & \textcolor[rgb]{ .682,  .667,  .667}{27.7\textsubscript{$\pm$0.9}} & \textcolor[rgb]{ .682,  .667,  .667}{35.7\textsubscript{$\pm$0.5}} & \textcolor[rgb]{ .682,  .667,  .667}{47.4\textsubscript{$\pm$1.1}} &      & \textcolor[rgb]{ .682,  .667,  .667}{60.3} & \textcolor[rgb]{ .682,  .667,  .667}{66.3} & \textcolor[rgb]{ .682,  .667,  .667}{70.5} \\
\bottomrule[1pt]
\end{tabular}%

  \label{tab:main_table}%

\end{table*}%

We conduct extensive node classification experiments to verify the effectiveness of our \model. We focus on the most prevalent downstream task, node classification. 
% Code \& data at \url{https://github.com/anonymous-596384/LHGNN} for review.

\subsection{Experimental Setup}

\paragraph{Datasets.} We evaluate the proposed framework on five popular benchmark datasets, including DBLP \cite{dblp}, Pubmed \cite{pubmed}, CoraFull \cite{corafull}, Coauthor-CS \cite{coauthorcs}, and ogbn-arxiv \cite{ogbnarxiv}. We provide further details in Appendix \sref{sec:appdendix_datasets}. 

\paragraph{Baselines.} We test the following pre-training strategies as baseline methods to initialize GNN models. For all methods, following GPPT \cite{sun2022gppt} we implement the downstream node classification as a similarity prediction problem described in \sref{sec:downstream} for a fair comparison.
\begin{itemize}[leftmargin=*] 
    \item \method{CL} \cite{you2020graph} uses constrastive learning for learning unsupervised representations. The augmentation (dropping, perturbation, and masking) ratio is set at 0.2 following the original setting.
    \item \method{GPT-GNN} \cite{hu2020gpt} is also a popular hybrid method, which integrates feature prediction and structure generation. We keep the same hyper-parameters as the original work.
    \item \method{GPPT} \cite{sun2022gppt} (or \method{Edge}) uses edge prediction as the pre-training task and reformulates the downstream node classification as an edge prediction problem. We implement the pretext task in \sref{sec:edge} which is naturally consistent with downstream task formulation.
\end{itemize}

Our \model~integrates \method{Edge} and \method{$k$-NN}, proposed in \sref{sec:reachability}. We further remove the prompt nodes to test a vanilla hybrid pre-training method, namely \method{Hybrid} (\method{Edge} + \method{$k$-NN}). We also test the single method \method{$k$-NN} for comparison.

\paragraph{Backbone.} We adopt the widely-used GNN -- Graph Attention Networks (GAT) \cite{veličković2018graph} as the main backbone. Training details can be found in Appendix~\sref{sec:appendix_training_details}.

\subsection{Experimental Results}

\label{sec:experiment_results}

The performance of downstream node classification tasks with different pre-training methods and few-shot settings is summarized in Table \ref{tab:main_table}. We report the performance of \model~by executing prompt transferability test on the \emph{validation set}. We also report the performance of two initialization methods as a reference, but do not count them as comparable methods for the fair comparison since it can be understood as performing prompt-based transferability test on the \emph{test set}. Particularly, we have the following findings.

\begin{itemize}[leftmargin=*] 
    \item \model~significantly enhances the performances. It achieves the best performance among all methods in 12 out of 15 settings. On average, \model~achieves relative performance gains of 1.9\%, 2.6\%, and 4.4\% over the best baseline \method{CL} in the 8-, 32- and 64-shot settings, respectively, which implies that \model~works better with larger datasets. 
    \item \model~outperforms \method{Edge}, \method{$k$-NN}, and \method{Hybrid} in 14 out of 15 settings, showing a strong integration capability for single tasks. Besides, \method{Hybrid} produces inferior performance compared to \method{Edge} and \method{$k$-NN} in 10 settings, which suggests that \emph{the performance of vanilla hybrid methods can be pulled down by poor tasks inside them due to the lack of useful task-specific knowledge}.
    \item \model~can find the best initialization task embedding in 11 settings (\ie, gets consistent results with the better one of \method{init with Edge} and \method{init with Reach}), demonstrating the effectiveness of the prompt transferability test even on a small validation set. In the 128-shot settings, it finds the best initialization method on all datasets successfully. This shows the scalability of the prompt transferability test, since as the amount of data increases, the distribution of the validation set will become closer and closer to the true distribution.
\end{itemize}

% Table generated by Excel2LaTeX from sheet 'graph'
\begin{table}[t]
  \centering
  \caption{Ablation study on three datasets in the 32-shot setting. Indentation means that the experimental conﬁguration is a modification based on the up-level indentation. }
    \begin{tabular}{lccc}
    \toprule[1pt]
    \multirow{2}[4]{*}{} & \textbf{DBLP} & \textbf{CoraFull} & \textbf{CS} \\
\cmidrule{2-4}         & \multicolumn{3}{c}{32-shot} \\
    \midrule[1pt]
    \model~(\method{Edge} + \method{$k$-NN}) & 73.6\textsubscript{$\pm$0.7} & 52.1\textsubscript{$\pm$1.7} & 89.5\textsubscript{$\pm$0.1} \\
     \quad- w/o position embedding & 72.8\textsubscript{$\pm$0.9} & 48.7\textsubscript{$\pm$0.3} & 89.5\textsubscript{$\pm$0.2} \\
     \quad\quad - init with \method{Edge} & \textcolor[rgb]{ .682,  .667,  .667}{72.6\textsubscript{$\pm$1.0}} & \textcolor[rgb]{ .682,  .667,  .667}{48.7\textsubscript{$\pm$0.3}} & \textcolor[rgb]{ .682,  .667,  .667}{89.5\textsubscript{$\pm$0.2}} \\
     \quad\quad - init with \method{$k$-NN} & \textcolor[rgb]{ .682,  .667,  .667}{72.8\textsubscript{$\pm$0.9}} & \textcolor[rgb]{ .682,  .667,  .667}{47.6\textsubscript{$\pm$1.0}} & \textcolor[rgb]{ .682,  .667,  .667}{88.5\textsubscript{$\pm$0.6}} \\
     \quad- w/o prompt & 70.4\textsubscript{$\pm$1.1} & 48.1\textsubscript{$\pm$1.1} & 85.6\textsubscript{$\pm$0.1} \\
    \bottomrule[1pt]
    \end{tabular}%
  \label{tab:ablation}%
\end{table}%

\stitle{Ablation studies on prompt node components. } We analyze the effectiveness of the two components in prompt nodes, position embedding and task embedding. We remove the task embedding and let the prompt node feature fill with task embedding only. Table \ref{tab:ablation} summarizes the results. We can find that even without position embedding, using task embedding to distinguish pre-train tasks outperforms the vanilla hybrid method. On average, the performance improvement brought by task embedding accounts for 61.6\% of the total improvement. This suggests that \emph{by injecting the task type information in the pre-training stage, the GNN model is able to capture the task-specific and task-independent knowledge separately, which is beneficial for learning generic structural and transferable semantics of the input graph.}

Moreover, compared with Table \ref{tab:main_table}, \model~with a pre-training task as initialization outperforms this task alone significantly. In the 32-shot setting on Coauthor-CS, \model~initialized with \method{Edge} achieves 4.6\% gains over \method{Edge}, and \model~initialized with \method{$k$-NN} achieves 3.4\% gains over \method{$k$-NN}. One possible reason is that \emph{hybrid training learns richer knowledge than the separated training}.

% Table generated by Excel2LaTeX from sheet 'graph'
\begin{table}[t]
  \centering
  \caption{Generalizability studies on more pre-training combinations. Bold: best results for each combination. }
    \begin{tabular}{lccc}
    \toprule[1pt]
         & \multicolumn{3}{c}{\textbf{Pubmed}} \\
\cmidrule{2-4}         & 8-shot & 32-shot & 128-shot \\
    \midrule[1pt]
    \method{Hybrid} (\method{CL} + \method{$k$-NN}) & 74.1\textsubscript{$\pm$0.9} & 79.4\textsubscript{$\pm$0.5} & 81.2\textsubscript{$\pm$0.3} \\
    \model~(\method{CL} + \method{$k$-NN}) & \boldmath{}\textbf{76.0\textsubscript{$\pm$0.7}}\unboldmath{} & \boldmath{}\textbf{80.3\textsubscript{$\pm$0.4}}\unboldmath{} & \boldmath{}\textbf{82.8\textsubscript{$\pm$0.2}}\unboldmath{} \\
      \quad- init with \method{CL} & \textcolor[rgb]{ .682,  .667,  .667}{75.9\textsubscript{$\pm$0.6}} & \textcolor[rgb]{ .682,  .667,  .667}{80.3\textsubscript{$\pm$0.4}} & \textcolor[rgb]{ .682,  .667,  .667}{82.8\textsubscript{$\pm$0.2}} \\
      \quad- init with \method{$k$-NN} & \textcolor[rgb]{ .682,  .667,  .667}{76.0\textsubscript{$\pm$0.7}} & \textcolor[rgb]{ .682,  .667,  .667}{80.7\textsubscript{$\pm$0.2}} & \textcolor[rgb]{ .682,  .667,  .667}{82.8\textsubscript{$\pm$0.2}} \\
    \midrule
    \method{Hybrid} (\method{CL} + \method{Edge}) & \boldmath{}\textbf{72.8\textsubscript{$\pm$0.8}}\unboldmath{} & 78.3\textsubscript{$\pm$0.4} & 81.4\textsubscript{$\pm$0.1} \\
    \model~(\method{CL} + \method{Edge}) & 71.1\textsubscript{$\pm$2.7} & \boldmath{}\textbf{79.4\textsubscript{$\pm$0.9}}\unboldmath{} & \boldmath{}\textbf{81.9\textsubscript{$\pm$0.3}}\unboldmath{} \\
      \quad- init with \method{CL} & \textcolor[rgb]{ .682,  .667,  .667}{70.5\textsubscript{$\pm$2.0}} & \textcolor[rgb]{ .682,  .667,  .667}{79.4\textsubscript{$\pm$0.9}} & \textcolor[rgb]{ .682,  .667,  .667}{81.9\textsubscript{$\pm$0.3}} \\
      \quad- init with \method{Edge} & \textcolor[rgb]{ .682,  .667,  .667}{71.1\textsubscript{$\pm$2.7}} & \textcolor[rgb]{ .682,  .667,  .667}{79.3\textsubscript{$\pm$0.6}} & \textcolor[rgb]{ .682,  .667,  .667}{81.8\textsubscript{$\pm$0.2}} \\
    \bottomrule[1pt]
    \end{tabular}%
  \label{tab:general_tasks}%
\end{table}%

\stitle{Generalizability studies on pre-training tasks. } Our proposed \model~is a unifying framework supporting arbitrary pre-training strategies. Considering \method{CL} is also a similarity prediction problem that is consistent with the training objective of the downstream task, it can be naturally incorporated into our framework. Therefore, we further evaluate the combination of \method{CL} + \method{$k$-NN} and \method{CL} + \method{Edge} as a generalizability test on Pubmed.

In Table \ref{tab:general_tasks}, we report the results compared to the vanilla hybrid training methods. Overall, \model~outperforms the \method{hybrid} for each combination in 5 out of 6 settings, which demonstrates the generalizability of our proposed \model~framework to various pre-training combinations. Particularly, compared to the respective \method{hybrid} baseline, \model~(\method{Edge} + \method{$k$-NN}) and \model~(\method{CL} + \method{$k$-NN}) bring greater improvements than \model~(\method{CL} + \method{Edge}). To explain, given that the \method{CL} task is similar to \method{Edge} since both of them are node-node level tasks, while \method{$k$-NN} is a node-group level task. Therefore, one possible explanation is that \emph{when the integrated pre-training tasks are more heterogeneous, \model~can learn more abundant and essential knowledge, which benefits downstream tasks}.

% Table generated by Excel2LaTeX from sheet 'graph'
\begin{table}[t]
  \centering
  \caption{Generalizability studies on other backbones. Bold: best results for each backbone. \underline{Underline}: best results among \method{Edge}-based and \method{$k$-NN}-based methods for each backbone.}
    \begin{tabular}{lccc}
    \toprule[1pt]
         & \multicolumn{3}{c}{\textbf{Pubmed}} \\
\cmidrule{2-4}         & 8-shot & 32-shot & 128-shot \\
    \midrule[1pt]
    \multicolumn{4}{c}{\textit{backbone: GCN}} \\
    \midrule
    \method{No Pre-train} & 69.7\textsubscript{$\pm$0.5} & 76.5\textsubscript{$\pm$0.3} & 80.4\textsubscript{$\pm$0.5} \\
    \method{CL} & \boldmath{}\textbf{74.7\textsubscript{$\pm$0.5}}\unboldmath{} & 79.1\textsubscript{$\pm$0.3} & 81.6\textsubscript{$\pm$0.2} \\
    \method{GPPT} (\method{Edge}) & 71.1\textsubscript{$\pm$0.9} & 77.7\textsubscript{$\pm$0.4} & 81.0\textsubscript{$\pm$0.3} \\
    \method{$k$-NN} & 74.5\textsubscript{$\pm$2.0} & 80.0\textsubscript{$\pm$0.4} & 82.0\textsubscript{$\pm$0.3} \\
    \method{Hybrid} (\method{Edge} + \method{$k$-NN}) & \underline{74.6\textsubscript{$\pm$0.7}} & 79.8\textsubscript{$\pm$0.1} & 81.7\textsubscript{$\pm$0.3} \\
    \midrule
    \model~(\method{Edge} + \method{$k$-NN}) & 74.3\textsubscript{$\pm$1.1} & \underline{\boldmath{}\textbf{80.6\textsubscript{$\pm$0.6}}\unboldmath{}} & \underline{\boldmath{}\textbf{83.5\textsubscript{$\pm$0.1}}\unboldmath{}} \\
     \quad- init with \method{Edge} & \textcolor[rgb]{ .682,  .667,  .667}{74.0\textsubscript{$\pm$0.5}} & \textcolor[rgb]{ .682,  .667,  .667}{80.4\textsubscript{$\pm$0.7}} & \textcolor[rgb]{ .682,  .667,  .667}{83.7\textsubscript{$\pm$0.2}} \\
     \quad- init with \method{$k$-NN} & \textcolor[rgb]{ .682,  .667,  .667}{74.3\textsubscript{$\pm$1.1}} & \textcolor[rgb]{ .682,  .667,  .667}{80.6\textsubscript{$\pm$0.6}} & \textcolor[rgb]{ .682,  .667,  .667}{83.5\textsubscript{$\pm$0.1}} \\
    \midrule[1pt]
    \multicolumn{4}{c}{\textit{backbone: SAGE}} \\
    \midrule
    \method{No Pre-train} & 68.0\textsubscript{$\pm$0.5} & 75.5\textsubscript{$\pm$0.6} & 80.6\textsubscript{$\pm$0.6} \\
    \method{CL} & 71.9\textsubscript{$\pm$2.0} & 79.1\textsubscript{$\pm$0.8} & 82.2\textsubscript{$\pm$0.5} \\
    \method{GPPT} (\method{Edge}) & 71.6\textsubscript{$\pm$0.8} & 78.4\textsubscript{$\pm$0.3} & 81.6\textsubscript{$\pm$0.3} \\
    \method{$k$-NN} & 75.3\textsubscript{$\pm$1.0} & 79.8\textsubscript{$\pm$0.4} & 81.9\textsubscript{$\pm$0.5} \\
    \method{Hybrid} (\method{Edge} + \method{$k$-NN}) & 75.7\textsubscript{$\pm$0.5} & 79.7\textsubscript{$\pm$0.2} & 83.1\textsubscript{$\pm$0.2} \\
    \midrule
    \model~(\method{Edge} + \method{$k$-NN}) & \underline{\boldmath{}\textbf{76.3\textsubscript{$\pm$0.3}}\unboldmath{}} & \underline{\boldmath{}\textbf{80.9\textsubscript{$\pm$0.1}}\unboldmath{}} & \underline{\boldmath{}\textbf{84.0\textsubscript{$\pm$0.2}}\unboldmath{}} \\
     \quad- init with \method{Edge} & \textcolor[rgb]{ .682,  .667,  .667}{76.0\textsubscript{$\pm$0.3}} & \textcolor[rgb]{ .682,  .667,  .667}{80.9\textsubscript{$\pm$0.2}} & \textcolor[rgb]{ .682,  .667,  .667}{84.0\textsubscript{$\pm$0.2}} \\
     \quad- init with \method{$k$-NN} & \textcolor[rgb]{ .682,  .667,  .667}{76.3\textsubscript{$\pm$0.3}} & \textcolor[rgb]{ .682,  .667,  .667}{80.9\textsubscript{$\pm$0.1}} & \textcolor[rgb]{ .682,  .667,  .667}{83.8\textsubscript{$\pm$0.2}} \\
    \bottomrule[1pt]
    \end{tabular}%
  \label{tab:general_backbone}%
\end{table}%

\stitle{Generalizability studies on the backbones. }  We investigate whether the other GNN architecture can benefit from the proposed pre-training framework. Therefore, we test GCN \cite{kipf2017semi} and SAGE \cite{hamilton2017inductive} as other backbones on Pubmed, in which all hyper-parameters remain the same. The results are reported in Table \ref{tab:general_backbone}. We can observe that the proposed pre-training framework can augment the downstream performance for all the GNN architectures, achieving the best performance in 5 out of 6 settings, which shows the generalizability to other backbones.

% Table generated by Excel2LaTeX from sheet 'graph'

\begin{table}[t]
  \centering
  \caption{Link prediction test on four datasets without fine-tuning. The evaluation metric is AUC (\%). \method{E}: \method{Edge}, $k$: \method{$k$-NN}.}
% Table generated by Excel2LaTeX from sheet 'graph'
\begin{tabular}{lccc}
\toprule[1pt]
     & \textbf{DBLP} & \textbf{CS} & \textbf{Pubmed} \\
\midrule[1pt]
\method{No pre-train} & 74.1\textsubscript{$\pm$0.6} &  88.9\textsubscript{$\pm$0.8} & 74.2\textsubscript{$\pm$1.7} \\
\method{Edge} & 92.0\textsubscript{$\pm$0.8} & 95.9\textsubscript{$\pm$0.3} & 93.7\textsubscript{$\pm$0.5} \\
\method{$k$-NN} & 84.8\textsubscript{$\pm$1.6} & 88.4\textsubscript{$\pm$0.4} & 87.1\textsubscript{$\pm$0.9} \\
\method{Hybrid} (\method{Edge} + \method{$k$-NN}) & 89.0\textsubscript{$\pm$0.7} & 92.7\textsubscript{$\pm$0.2} & 91.5\textsubscript{$\pm$0.8} \\
\midrule
\model~(\method{Edge} + \method{$k$-NN}) & & &  \\
   \quad- init with \method{Edge} & 90.4\textsubscript{$\pm$0.2} & 92.8\textsubscript{$\pm$0.4} & 91.6\textsubscript{$\pm$2.4} \\
  \quad- init with \method{$k$-NN} & 89.3\textsubscript{$\pm$0.7} & 90.7\textsubscript{$\pm$1.5} & 79.4\textsubscript{$\pm$7.3} \\
\bottomrule[1pt]
\end{tabular}%

  \label{tab:link}%
\end{table}%

\stitle{Deep into the role of task embedding. } \model~performs well in the node classification task since it captures all aspects of semantic information by integrating multiple pre-training tasks. Furthermore, we investigate how the mutual interference of these tasks inside the model change by introducing distinguishable task embeddings. Specifically, we test the \textit{link prediction} ability of the models pre-trained with hybrid tasks, since \method{Edge} is one of these tasks and we want to investigate how much it will be interfered with by the other task, \method{$k$-NN}. To focus on the role of task embedding, we remove the position embedding in our proposed \model.

Table \ref{tab:link} summarizes the results. We report the performance of models pre-trained only with the \method{Edge} task as an upper bound. One can find that 1) \model~initialized with \method{Edge} task embedding outperforms the one initialized with \method{$k$-NN} significantly especially on Pubmed, showing that \emph{the task embedding stores the knowledge of the corresponding task and facilitates the GNN to memorize it}; 2) compared to the vanilla hybrid method, \model~achieves better performance consistently, implying that \emph{task embedding reduces the mutual interference between pre-training tasks}; and 3) \method{Edge} still achieves the best link prediction performance. This is reasonable since in \method{Edge} all of the parameters of GNN are trained for link prediction, while in \model~the parameters are trained for multiple tasks. This suggests that \emph{there is still room for task embedding to store richer task-specific knowledge}.

\begin{figure}[t]
    \centering
    \subfigure[Number of anchors $\nicefrac{m}{|\mathcal{V}|}$ (\%)]{
        \includegraphics[width=0.46\linewidth]{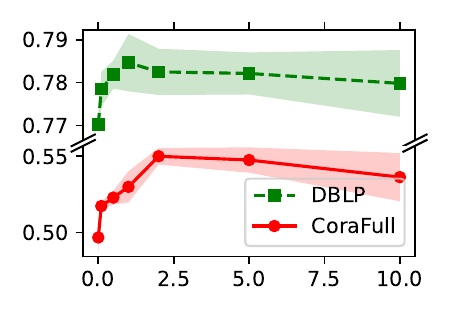}
        % \vspace{-2mm}
        \label{fig:params_anchors}
    }
    \subfigure[Position encoding step $t$]{
        \includegraphics[width=0.46\linewidth]{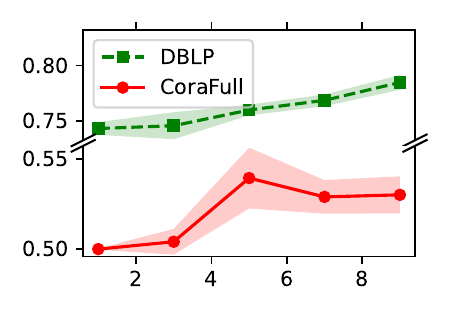}
        % \vspace{-2mm}
        \label{fig:params_pos_step}
    }
    \subfigure[Position embedding weight $w^{\text{pos}}$]{
        \includegraphics[width=0.46\linewidth]{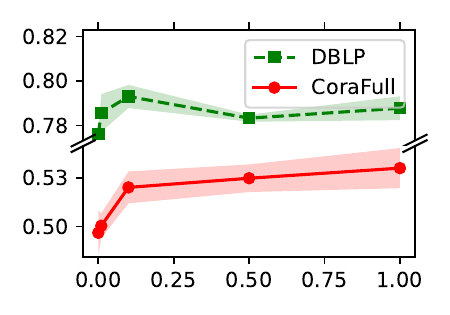}
        % \vspace{-2mm}
        \label{fig:params_pos_weight}
    }
    \subfigure[$k$-NN reachability step $t'$]{
        \includegraphics[width=0.46\linewidth]{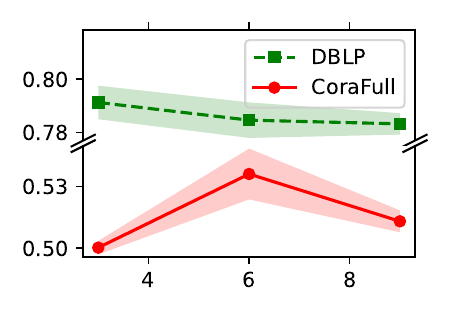}
        % \vspace{-2mm}
        \label{fig:params_knn_step}
    }
    \vspace{-3mm}
    \caption{Parameters sensitivity. The evaluation metric is Micro-F1. The standard deviation is displayed with the shadow areas.}
    \label{fig:parameters}
    % \vspace{-3mm}
\end{figure}

\stitle{}

\stitle{Parameters Sensitivity.}
We evaluate the sensitivity of several important hyper-parameters in \model, and show their impact in Figure \ref{fig:parameters}. We can observe that 1) for the number of anchors $m$ used in position embedding, too small values make the position estimate inaccurate, while too large values increase the parameter size and may result in overfitting. Moderate values such as 1\% or 2\% of the total nodes are optimal; 2) the optimal random walk step $t$ used in position embedding depends on the graph topology since the minimum number of steps to traverse the whole graph is diverse on different datasets. Five steps are enough in CoraFull, while DBLP needs more; 3) for the weight of position embedding $w^{\text{pos}}$, choosing relatively large values such as $0.1$ or more can make position embedding's prompt effect stronger and benefit the performance; and 4) the optimal random walk step $t$ used in $k$-NN similarity prediction task is about 6.
\section{Conclusion and Future Work}

In this paper, we investigate how to bridge the semantic gap between graph pre-training and downstream tasks. We propose a unifying framework \model~for graph hybrid pre-training by introducing prompt nodes without modifying the model structure, to incorporate task type and position information which traditional methods miss. Based on it, we propose a prompt-based transferability test to reduce the semantic gap. We also propose a novel pre-training task, $k$-NN similarity prediction, combined with the classical edge prediction for hybrid pre-training. Extensive experiments on five benchmark datasets demonstrate the effectiveness of our proposed \model.

\stitle{Future work.} Although unifying the task objectives is beneficial, it is non-trivial to reformulate some tasks as a similarity prediction task, \eg, attribute masking \cite{Hu2020Strategies} pre-training task and node regression downstream task. 
However, we can still utilize \model~framework to bridge the semantic gap, \eg, attaching the prompt to the node whose attributes are masked. 
By prompting GNNs with the task type, GNNs can learn heterogeneous knowledge without being influenced by other tasks.
We leave incorporating more pre-training tasks into our \model~framework as future work.

\newpage

%%
%% The acknowledgments section is defined using the "acks" environment
%% (and NOT an unnumbered section). This ensures the proper
%% identification of the section in the article metadata, and the
%% consistent spelling of the heading.

%%
%% The next two lines define the bibliography style to be used, and
%% the bibliography file.
\bibliographystyle{ACM-Reference-Format}
\bibliography{reference}

%%% -*-BibTeX-*-
%%% Do NOT edit. File created by BibTeX with style
%%% ACM-Reference-Format-Journals [18-Jan-2012].

\begin{thebibliography}{47}

%%% ====================================================================
%%% NOTE TO THE USER: you can override these defaults by providing
%%% customized versions of any of these macros before the \bibliography
%%% command.  Each of them MUST provide its own final punctuation,
%%% except for \shownote{}, \showDOI{}, and \showURL{}.  The latter two
%%% do not use final punctuation, in order to avoid confusing it with
%%% the Web address.
%%%
%%% To suppress output of a particular field, define its macro to expand
%%% to an empty string, or better, \unskip, like this:
%%%
%%% \newcommand{\showDOI}[1]{\unskip}   % LaTeX syntax
%%%
%%% \def \showDOI #1{\unskip}           % plain TeX syntax
%%%
%%% ====================================================================

\ifx \showCODEN    \undefined \def \showCODEN     #1{\unskip}     \fi
\ifx \showDOI      \undefined \def \showDOI       #1{#1}\fi
\ifx \showISBNx    \undefined \def \showISBNx     #1{\unskip}     \fi
\ifx \showISBNxiii \undefined \def \showISBNxiii  #1{\unskip}     \fi
\ifx \showISSN     \undefined \def \showISSN      #1{\unskip}     \fi
\ifx \showLCCN     \undefined \def \showLCCN      #1{\unskip}     \fi
\ifx \shownote     \undefined \def \shownote      #1{#1}          \fi
\ifx \showarticletitle \undefined \def \showarticletitle #1{#1}   \fi
\ifx \showURL      \undefined \def \showURL       {\relax}        \fi
% The following commands are used for tagged output and should be
% invisible to TeX
\providecommand\bibfield[2]{#2}
\providecommand\bibinfo[2]{#2}
\providecommand\natexlab[1]{#1}
\providecommand\showeprint[2][]{arXiv:#2}

\bibitem[Bojchevski and Günnemann(2018)]%
        {corafull}
\bibfield{author}{\bibinfo{person}{Aleksandar Bojchevski} {and}
  \bibinfo{person}{Stephan Günnemann}.} \bibinfo{year}{2018}\natexlab{}.
\newblock \showarticletitle{Deep Gaussian Embedding of Graphs: Unsupervised
  Inductive Learning via Ranking}. In \bibinfo{booktitle}{\emph{ICLR}}.
\newblock


\bibitem[Bruna et~al\mbox{.}(2013)]%
        {bruna2013spectral}
\bibfield{author}{\bibinfo{person}{Joan Bruna}, \bibinfo{person}{Wojciech
  Zaremba}, \bibinfo{person}{Arthur Szlam}, {and} \bibinfo{person}{Yann
  LeCun}.} \bibinfo{year}{2013}\natexlab{}.
\newblock \showarticletitle{Spectral networks and locally connected networks on
  graphs}.
\newblock \bibinfo{journal}{\emph{arXiv preprint arXiv:1312.6203}}
  (\bibinfo{year}{2013}).
\newblock


\bibitem[Dwivedi and Bresson(2020)]%
        {dwivedi2020generalization}
\bibfield{author}{\bibinfo{person}{Vijay~Prakash Dwivedi} {and}
  \bibinfo{person}{Xavier Bresson}.} \bibinfo{year}{2020}\natexlab{}.
\newblock \showarticletitle{A generalization of transformer networks to
  graphs}.
\newblock \bibinfo{journal}{\emph{arXiv preprint arXiv:2012.09699}}
  (\bibinfo{year}{2020}).
\newblock


\bibitem[Gao et~al\mbox{.}(2021)]%
        {gao-etal-2021-making}
\bibfield{author}{\bibinfo{person}{Tianyu Gao}, \bibinfo{person}{Adam Fisch},
  {and} \bibinfo{person}{Danqi Chen}.} \bibinfo{year}{2021}\natexlab{}.
\newblock \showarticletitle{Making Pre-trained Language Models Better Few-shot
  Learners}. In \bibinfo{booktitle}{\emph{ACL}}.
  \bibinfo{publisher}{Association for Computational Linguistics},
  \bibinfo{address}{Online}, \bibinfo{pages}{3816--3830}.
\newblock


\bibitem[Gilmer et~al\mbox{.}(2017)]%
        {gilmer2017neural}
\bibfield{author}{\bibinfo{person}{Justin Gilmer}, \bibinfo{person}{Samuel~S
  Schoenholz}, \bibinfo{person}{Patrick~F Riley}, \bibinfo{person}{Oriol
  Vinyals}, {and} \bibinfo{person}{George~E Dahl}.}
  \bibinfo{year}{2017}\natexlab{}.
\newblock \showarticletitle{Neural message passing for quantum chemistry}. In
  \bibinfo{booktitle}{\emph{ICML}}. PMLR, \bibinfo{pages}{1263--1272}.
\newblock


\bibitem[Gu et~al\mbox{.}(2022)]%
        {gu-etal-2022-ppt}
\bibfield{author}{\bibinfo{person}{Yuxian Gu}, \bibinfo{person}{Xu Han},
  \bibinfo{person}{Zhiyuan Liu}, {and} \bibinfo{person}{Minlie Huang}.}
  \bibinfo{year}{2022}\natexlab{}.
\newblock \showarticletitle{{PPT}: Pre-trained Prompt Tuning for Few-shot
  Learning}. In \bibinfo{booktitle}{\emph{ACL}}.
  \bibinfo{publisher}{Association for Computational Linguistics},
  \bibinfo{address}{Dublin, Ireland}, \bibinfo{pages}{8410--8423}.
\newblock


\bibitem[Hamilton et~al\mbox{.}(2017)]%
        {hamilton2017inductive}
\bibfield{author}{\bibinfo{person}{Will Hamilton}, \bibinfo{person}{Zhitao
  Ying}, {and} \bibinfo{person}{Jure Leskovec}.}
  \bibinfo{year}{2017}\natexlab{}.
\newblock \showarticletitle{Inductive representation learning on large graphs}.
\newblock \bibinfo{journal}{\emph{NeurIPS}}  \bibinfo{volume}{30}
  (\bibinfo{year}{2017}).
\newblock


\bibitem[Hu et~al\mbox{.}(2020b)]%
        {ogbnarxiv}
\bibfield{author}{\bibinfo{person}{Weihua Hu}, \bibinfo{person}{Matthias Fey},
  \bibinfo{person}{Marinka Zitnik}, \bibinfo{person}{Yuxiao Dong},
  \bibinfo{person}{Hongyu Ren}, \bibinfo{person}{Bowen Liu},
  \bibinfo{person}{Michele Catasta}, {and} \bibinfo{person}{Jure Leskovec}.}
  \bibinfo{year}{2020}\natexlab{b}.
\newblock \showarticletitle{Open graph benchmark: Datasets for machine learning
  on graphs}.
\newblock \bibinfo{journal}{\emph{NeurIPS}}  \bibinfo{volume}{33}
  (\bibinfo{year}{2020}), \bibinfo{pages}{22118--22133}.
\newblock


\bibitem[Hu* et~al\mbox{.}(2020)]%
        {Hu2020Strategies}
\bibfield{author}{\bibinfo{person}{Weihua Hu*}, \bibinfo{person}{Bowen Liu*},
  \bibinfo{person}{Joseph Gomes}, \bibinfo{person}{Marinka Zitnik},
  \bibinfo{person}{Percy Liang}, \bibinfo{person}{Vijay Pande}, {and}
  \bibinfo{person}{Jure Leskovec}.} \bibinfo{year}{2020}\natexlab{}.
\newblock \showarticletitle{Strategies for Pre-training Graph Neural Networks}.
  In \bibinfo{booktitle}{\emph{ICLR}}.
\newblock


\bibitem[Hu et~al\mbox{.}(2020a)]%
        {hu2020gpt}
\bibfield{author}{\bibinfo{person}{Ziniu Hu}, \bibinfo{person}{Yuxiao Dong},
  \bibinfo{person}{Kuansan Wang}, \bibinfo{person}{Kai-Wei Chang}, {and}
  \bibinfo{person}{Yizhou Sun}.} \bibinfo{year}{2020}\natexlab{a}.
\newblock \showarticletitle{GPT-GNN: Generative Pre-Training of Graph Neural
  Networks}. In \bibinfo{booktitle}{\emph{KDD}} (Virtual Event, CA, USA)
  \emph{(\bibinfo{series}{KDD '20})}. \bibinfo{publisher}{Association for
  Computing Machinery}, \bibinfo{address}{New York, NY, USA},
  \bibinfo{pages}{1857–1867}.
\newblock
\showISBNx{9781450379984}


\bibitem[Jin et~al\mbox{.}(2020)]%
        {jin2020self}
\bibfield{author}{\bibinfo{person}{Wei Jin}, \bibinfo{person}{Tyler Derr},
  \bibinfo{person}{Haochen Liu}, \bibinfo{person}{Yiqi Wang},
  \bibinfo{person}{Suhang Wang}, \bibinfo{person}{Zitao Liu}, {and}
  \bibinfo{person}{Jiliang Tang}.} \bibinfo{year}{2020}\natexlab{}.
\newblock \showarticletitle{Self-supervised learning on graphs: Deep insights
  and new direction}.
\newblock \bibinfo{journal}{\emph{arXiv preprint arXiv:2006.10141}}
  (\bibinfo{year}{2020}).
\newblock


\bibitem[Kipf and Welling(2016)]%
        {kipf2016variational}
\bibfield{author}{\bibinfo{person}{Thomas~N Kipf} {and} \bibinfo{person}{Max
  Welling}.} \bibinfo{year}{2016}\natexlab{}.
\newblock \showarticletitle{Variational Graph Auto-Encoders}.
\newblock \bibinfo{journal}{\emph{NIPS Workshop on Bayesian Deep Learning}}
  (\bibinfo{year}{2016}).
\newblock


\bibitem[Kipf and Welling(2017)]%
        {kipf2017semi}
\bibfield{author}{\bibinfo{person}{Thomas~N. Kipf} {and} \bibinfo{person}{Max
  Welling}.} \bibinfo{year}{2017}\natexlab{}.
\newblock \showarticletitle{Semi-Supervised Classification with Graph
  Convolutional Networks}. In \bibinfo{booktitle}{\emph{ICLR}}.
\newblock


\bibitem[Lester et~al\mbox{.}(2021)]%
        {lester-etal-2021-power}
\bibfield{author}{\bibinfo{person}{Brian Lester}, \bibinfo{person}{Rami
  Al-Rfou}, {and} \bibinfo{person}{Noah Constant}.}
  \bibinfo{year}{2021}\natexlab{}.
\newblock \showarticletitle{The Power of Scale for Parameter-Efficient Prompt
  Tuning}. In \bibinfo{booktitle}{\emph{EMNLP}}.
  \bibinfo{publisher}{Association for Computational Linguistics},
  \bibinfo{address}{Online and Punta Cana, Dominican Republic},
  \bibinfo{pages}{3045--3059}.
\newblock


\bibitem[Li and Liang(2021)]%
        {li-liang-2021-prefix}
\bibfield{author}{\bibinfo{person}{Xiang~Lisa Li} {and} \bibinfo{person}{Percy
  Liang}.} \bibinfo{year}{2021}\natexlab{}.
\newblock \showarticletitle{Prefix-Tuning: Optimizing Continuous Prompts for
  Generation}. In \bibinfo{booktitle}{\emph{ACL}}.
  \bibinfo{publisher}{Association for Computational Linguistics},
  \bibinfo{address}{Online}, \bibinfo{pages}{4582--4597}.
\newblock


\bibitem[Liu et~al\mbox{.}(2021)]%
        {liu2021gpt}
\bibfield{author}{\bibinfo{person}{Xiao Liu}, \bibinfo{person}{Yanan Zheng},
  \bibinfo{person}{Zhengxiao Du}, \bibinfo{person}{Ming Ding},
  \bibinfo{person}{Yujie Qian}, \bibinfo{person}{Zhilin Yang}, {and}
  \bibinfo{person}{Jie Tang}.} \bibinfo{year}{2021}\natexlab{}.
\newblock \showarticletitle{GPT understands, too}.
\newblock \bibinfo{journal}{\emph{arXiv preprint arXiv:2103.10385}}
  (\bibinfo{year}{2021}).
\newblock


\bibitem[Liu et~al\mbox{.}(2022)]%
        {liu2022graph}
\bibfield{author}{\bibinfo{person}{Yixin Liu}, \bibinfo{person}{Ming Jin},
  \bibinfo{person}{Shirui Pan}, \bibinfo{person}{Chuan Zhou},
  \bibinfo{person}{Yu Zheng}, \bibinfo{person}{Feng Xia}, {and}
  \bibinfo{person}{Philip Yu}.} \bibinfo{year}{2022}\natexlab{}.
\newblock \showarticletitle{Graph self-supervised learning: A survey}.
\newblock \bibinfo{journal}{\emph{TKDE}} (\bibinfo{year}{2022}).
\newblock


\bibitem[Loshchilov and Hutter(2019)]%
        {loshchilov2018decoupled}
\bibfield{author}{\bibinfo{person}{Ilya Loshchilov} {and}
  \bibinfo{person}{Frank Hutter}.} \bibinfo{year}{2019}\natexlab{}.
\newblock \showarticletitle{Decoupled Weight Decay Regularization}. In
  \bibinfo{booktitle}{\emph{ICLR}}.
\newblock


\bibitem[Lu et~al\mbox{.}(2021)]%
        {lu2021learning}
\bibfield{author}{\bibinfo{person}{Yuanfu Lu}, \bibinfo{person}{Xunqiang
  Jiang}, \bibinfo{person}{Yuan Fang}, {and} \bibinfo{person}{Chuan Shi}.}
  \bibinfo{year}{2021}\natexlab{}.
\newblock \showarticletitle{Learning to pre-train graph neural networks}. In
  \bibinfo{booktitle}{\emph{AAAI}}, Vol.~\bibinfo{volume}{35}.
  \bibinfo{pages}{4276--4284}.
\newblock


\bibitem[Manessi and Rozza(2021)]%
        {manessi2021graph}
\bibfield{author}{\bibinfo{person}{Franco Manessi} {and}
  \bibinfo{person}{Alessandro Rozza}.} \bibinfo{year}{2021}\natexlab{}.
\newblock \showarticletitle{Graph-based neural network models with multiple
  self-supervised auxiliary tasks}.
\newblock \bibinfo{journal}{\emph{Pattern Recognition Letters}}
  \bibinfo{volume}{148} (\bibinfo{year}{2021}), \bibinfo{pages}{15--21}.
\newblock


\bibitem[Nishad et~al\mbox{.}(2021)]%
        {sunil2021graphreach}
\bibfield{author}{\bibinfo{person}{Sunil Nishad}, \bibinfo{person}{Shubhangi
  Agarwal}, \bibinfo{person}{Arnab Bhattacharya}, {and} \bibinfo{person}{Sayan
  Ranu}.} \bibinfo{year}{2021}\natexlab{}.
\newblock \showarticletitle{GraphReach: Position-Aware Graph Neural Network
  using Reachability Estimations}. In \bibinfo{booktitle}{\emph{IJCAI}},
  \bibfield{editor}{\bibinfo{person}{Zhi-Hua Zhou}} (Ed.).
  \bibinfo{publisher}{International Joint Conferences on Artificial
  Intelligence Organization}, \bibinfo{pages}{1527--1533}.
\newblock
\newblock
\shownote{Main Track}.


\bibitem[Oh~Song et~al\mbox{.}(2016)]%
        {oh2016deep}
\bibfield{author}{\bibinfo{person}{Hyun Oh~Song}, \bibinfo{person}{Yu Xiang},
  \bibinfo{person}{Stefanie Jegelka}, {and} \bibinfo{person}{Silvio Savarese}.}
  \bibinfo{year}{2016}\natexlab{}.
\newblock \showarticletitle{Deep metric learning via lifted structured feature
  embedding}. In \bibinfo{booktitle}{\emph{CVPR}}. \bibinfo{pages}{4004--4012}.
\newblock


\bibitem[Oord et~al\mbox{.}(2018)]%
        {oord2018representation}
\bibfield{author}{\bibinfo{person}{Aaron van~den Oord}, \bibinfo{person}{Yazhe
  Li}, {and} \bibinfo{person}{Oriol Vinyals}.} \bibinfo{year}{2018}\natexlab{}.
\newblock \showarticletitle{Representation learning with contrastive predictive
  coding}.
\newblock \bibinfo{journal}{\emph{arXiv preprint arXiv:1807.03748}}
  (\bibinfo{year}{2018}).
\newblock


\bibitem[Panagopoulos et~al\mbox{.}(2021)]%
        {panagopoulos2021transfer}
\bibfield{author}{\bibinfo{person}{George Panagopoulos},
  \bibinfo{person}{Giannis Nikolentzos}, {and} \bibinfo{person}{Michalis
  Vazirgiannis}.} \bibinfo{year}{2021}\natexlab{}.
\newblock \showarticletitle{Transfer graph neural networks for pandemic
  forecasting}. In \bibinfo{booktitle}{\emph{AAAI}}, Vol.~\bibinfo{volume}{35}.
  \bibinfo{pages}{4838--4845}.
\newblock


\bibitem[Petroni et~al\mbox{.}(2019)]%
        {petroni-etal-2019-language}
\bibfield{author}{\bibinfo{person}{Fabio Petroni}, \bibinfo{person}{Tim
  Rockt{\"a}schel}, \bibinfo{person}{Sebastian Riedel},
  \bibinfo{person}{Patrick Lewis}, \bibinfo{person}{Anton Bakhtin},
  \bibinfo{person}{Yuxiang Wu}, {and} \bibinfo{person}{Alexander Miller}.}
  \bibinfo{year}{2019}\natexlab{}.
\newblock \showarticletitle{Language Models as Knowledge Bases?}. In
  \bibinfo{booktitle}{\emph{EMNLP}}. \bibinfo{publisher}{Association for
  Computational Linguistics}, \bibinfo{address}{Hong Kong, China},
  \bibinfo{pages}{2463--2473}.
\newblock


\bibitem[Qiu et~al\mbox{.}(2020)]%
        {qiu2020gcc}
\bibfield{author}{\bibinfo{person}{Jiezhong Qiu}, \bibinfo{person}{Qibin Chen},
  \bibinfo{person}{Yuxiao Dong}, \bibinfo{person}{Jing Zhang},
  \bibinfo{person}{Hongxia Yang}, \bibinfo{person}{Ming Ding},
  \bibinfo{person}{Kuansan Wang}, {and} \bibinfo{person}{Jie Tang}.}
  \bibinfo{year}{2020}\natexlab{}.
\newblock \showarticletitle{Gcc: Graph contrastive coding for graph neural
  network pre-training}. In \bibinfo{booktitle}{\emph{KDD}}.
  \bibinfo{pages}{1150--1160}.
\newblock


\bibitem[Ramp{\'a}{\v{s}}ek et~al\mbox{.}(2022)]%
        {rampavsek2022recipe}
\bibfield{author}{\bibinfo{person}{Ladislav Ramp{\'a}{\v{s}}ek},
  \bibinfo{person}{Mikhail Galkin}, \bibinfo{person}{Vijay~Prakash Dwivedi},
  \bibinfo{person}{Anh~Tuan Luu}, \bibinfo{person}{Guy Wolf}, {and}
  \bibinfo{person}{Dominique Beaini}.} \bibinfo{year}{2022}\natexlab{}.
\newblock \showarticletitle{Recipe for a General, Powerful, Scalable Graph
  Transformer}.
\newblock \bibinfo{journal}{\emph{arXiv preprint arXiv:2205.12454}}
  (\bibinfo{year}{2022}).
\newblock


\bibitem[Rosenstein(2005)]%
        {Rosenstein2005ToTO}
\bibfield{author}{\bibinfo{person}{Michael~T. Rosenstein}.}
  \bibinfo{year}{2005}\natexlab{}.
\newblock \showarticletitle{To transfer or not to transfer}. In
  \bibinfo{booktitle}{\emph{NeurIPS}}.
\newblock


\bibitem[Schick and Sch{\"u}tze(2021)]%
        {schick-schutze-2021-exploiting}
\bibfield{author}{\bibinfo{person}{Timo Schick} {and} \bibinfo{person}{Hinrich
  Sch{\"u}tze}.} \bibinfo{year}{2021}\natexlab{}.
\newblock \showarticletitle{Exploiting Cloze-Questions for Few-Shot Text
  Classification and Natural Language Inference}. In
  \bibinfo{booktitle}{\emph{EACL}}. \bibinfo{publisher}{Association for
  Computational Linguistics}, \bibinfo{address}{Online},
  \bibinfo{pages}{255--269}.
\newblock


\bibitem[Schroff et~al\mbox{.}(2015)]%
        {schroff2015facenet}
\bibfield{author}{\bibinfo{person}{Florian Schroff}, \bibinfo{person}{Dmitry
  Kalenichenko}, {and} \bibinfo{person}{James Philbin}.}
  \bibinfo{year}{2015}\natexlab{}.
\newblock \showarticletitle{Facenet: A unified embedding for face recognition
  and clustering}. In \bibinfo{booktitle}{\emph{CVPR}}.
  \bibinfo{pages}{815--823}.
\newblock


\bibitem[Sen et~al\mbox{.}(2008)]%
        {pubmed}
\bibfield{author}{\bibinfo{person}{Prithviraj Sen}, \bibinfo{person}{Galileo
  Namata}, \bibinfo{person}{Mustafa Bilgic}, \bibinfo{person}{Lise Getoor},
  \bibinfo{person}{Brian Galligher}, {and} \bibinfo{person}{Tina Eliassi-Rad}.}
  \bibinfo{year}{2008}\natexlab{}.
\newblock \showarticletitle{Collective classification in network data}.
\newblock \bibinfo{journal}{\emph{AI magazine}} \bibinfo{volume}{29},
  \bibinfo{number}{3} (\bibinfo{year}{2008}), \bibinfo{pages}{93--93}.
\newblock


\bibitem[Shchur et~al\mbox{.}(2018)]%
        {coauthorcs}
\bibfield{author}{\bibinfo{person}{Oleksandr Shchur},
  \bibinfo{person}{Maximilian Mumme}, \bibinfo{person}{Aleksandar Bojchevski},
  {and} \bibinfo{person}{Stephan G{\"u}nnemann}.}
  \bibinfo{year}{2018}\natexlab{}.
\newblock \showarticletitle{Pitfalls of graph neural network evaluation}.
\newblock \bibinfo{journal}{\emph{arXiv preprint arXiv:1811.05868}}
  (\bibinfo{year}{2018}).
\newblock


\bibitem[Sun et~al\mbox{.}(2022)]%
        {sun2022gppt}
\bibfield{author}{\bibinfo{person}{Mingchen Sun}, \bibinfo{person}{Kaixiong
  Zhou}, \bibinfo{person}{Xin He}, \bibinfo{person}{Ying Wang}, {and}
  \bibinfo{person}{Xin Wang}.} \bibinfo{year}{2022}\natexlab{}.
\newblock \showarticletitle{GPPT: Graph Pre-Training and Prompt Tuning to
  Generalize Graph Neural Networks}. In \bibinfo{booktitle}{\emph{KDD}}
  (Washington DC, USA) \emph{(\bibinfo{series}{KDD '22})}.
  \bibinfo{publisher}{Association for Computing Machinery},
  \bibinfo{address}{New York, NY, USA}, \bibinfo{pages}{1717–1727}.
\newblock
\showISBNx{9781450393850}


\bibitem[Tang et~al\mbox{.}(2008)]%
        {dblp}
\bibfield{author}{\bibinfo{person}{Jie Tang}, \bibinfo{person}{Jing Zhang},
  \bibinfo{person}{Limin Yao}, \bibinfo{person}{Juanzi Li}, \bibinfo{person}{Li
  Zhang}, {and} \bibinfo{person}{Zhong Su}.} \bibinfo{year}{2008}\natexlab{}.
\newblock \showarticletitle{Arnetminer: extraction and mining of academic
  social networks}. In \bibinfo{booktitle}{\emph{KDD}}.
  \bibinfo{pages}{990--998}.
\newblock


\bibitem[Veličković et~al\mbox{.}(2018)]%
        {veličković2018graph}
\bibfield{author}{\bibinfo{person}{Petar Veličković},
  \bibinfo{person}{Guillem Cucurull}, \bibinfo{person}{Arantxa Casanova},
  \bibinfo{person}{Adriana Romero}, \bibinfo{person}{Pietro Liò}, {and}
  \bibinfo{person}{Yoshua Bengio}.} \bibinfo{year}{2018}\natexlab{}.
\newblock \showarticletitle{Graph Attention Networks}. In
  \bibinfo{booktitle}{\emph{ICLR}}.
\newblock


\bibitem[Vu et~al\mbox{.}(2022)]%
        {vu-etal-2022-spot}
\bibfield{author}{\bibinfo{person}{Tu Vu}, \bibinfo{person}{Brian Lester},
  \bibinfo{person}{Noah Constant}, \bibinfo{person}{Rami Al-Rfou{'}}, {and}
  \bibinfo{person}{Daniel Cer}.} \bibinfo{year}{2022}\natexlab{}.
\newblock \showarticletitle{{SP}o{T}: Better Frozen Model Adaptation through
  Soft Prompt Transfer}. In \bibinfo{booktitle}{\emph{ACL}}.
  \bibinfo{publisher}{Association for Computational Linguistics},
  \bibinfo{address}{Dublin, Ireland}, \bibinfo{pages}{5039--5059}.
\newblock


\bibitem[Wen et~al\mbox{.}(2016)]%
        {wen2016discriminative}
\bibfield{author}{\bibinfo{person}{Yandong Wen}, \bibinfo{person}{Kaipeng
  Zhang}, \bibinfo{person}{Zhifeng Li}, {and} \bibinfo{person}{Yu Qiao}.}
  \bibinfo{year}{2016}\natexlab{}.
\newblock \showarticletitle{A discriminative feature learning approach for deep
  face recognition}. In \bibinfo{booktitle}{\emph{ECCV}}. Springer,
  \bibinfo{pages}{499--515}.
\newblock


\bibitem[Wu et~al\mbox{.}(2020)]%
        {wu2020comprehensive}
\bibfield{author}{\bibinfo{person}{Zonghan Wu}, \bibinfo{person}{Shirui Pan},
  \bibinfo{person}{Fengwen Chen}, \bibinfo{person}{Guodong Long},
  \bibinfo{person}{Chengqi Zhang}, {and} \bibinfo{person}{S~Yu Philip}.}
  \bibinfo{year}{2020}\natexlab{}.
\newblock \showarticletitle{A comprehensive survey on graph neural networks}.
\newblock \bibinfo{journal}{\emph{IEEE TNNLS}} \bibinfo{volume}{32},
  \bibinfo{number}{1} (\bibinfo{year}{2020}), \bibinfo{pages}{4--24}.
\newblock


\bibitem[Xu et~al\mbox{.}(2020)]%
        {Xu2020Graph}
\bibfield{author}{\bibinfo{person}{Chunyan Xu}, \bibinfo{person}{Zhen Cui},
  \bibinfo{person}{Xiaobin Hong}, \bibinfo{person}{Tong Zhang},
  \bibinfo{person}{Jian Yang}, {and} \bibinfo{person}{Wei Liu}.}
  \bibinfo{year}{2020}\natexlab{}.
\newblock \showarticletitle{Graph inference learning for semi-supervised
  classification}. In \bibinfo{booktitle}{\emph{ICLR}}.
\newblock


\bibitem[Xu et~al\mbox{.}(2019)]%
        {xu2018how}
\bibfield{author}{\bibinfo{person}{Keyulu Xu}, \bibinfo{person}{Weihua Hu},
  \bibinfo{person}{Jure Leskovec}, {and} \bibinfo{person}{Stefanie Jegelka}.}
  \bibinfo{year}{2019}\natexlab{}.
\newblock \showarticletitle{How Powerful are Graph Neural Networks?}. In
  \bibinfo{booktitle}{\emph{ICLR}}.
\newblock


\bibitem[Ying et~al\mbox{.}(2021)]%
        {ying2021transformers}
\bibfield{author}{\bibinfo{person}{Chengxuan Ying}, \bibinfo{person}{Tianle
  Cai}, \bibinfo{person}{Shengjie Luo}, \bibinfo{person}{Shuxin Zheng},
  \bibinfo{person}{Guolin Ke}, \bibinfo{person}{Di He},
  \bibinfo{person}{Yanming Shen}, {and} \bibinfo{person}{Tie-Yan Liu}.}
  \bibinfo{year}{2021}\natexlab{}.
\newblock \showarticletitle{Do transformers really perform badly for graph
  representation?}
\newblock \bibinfo{journal}{\emph{NeurIPS}}  \bibinfo{volume}{34}
  (\bibinfo{year}{2021}), \bibinfo{pages}{28877--28888}.
\newblock


\bibitem[Ying et~al\mbox{.}(2018)]%
        {ying2018graph}
\bibfield{author}{\bibinfo{person}{Rex Ying}, \bibinfo{person}{Ruining He},
  \bibinfo{person}{Kaifeng Chen}, \bibinfo{person}{Pong Eksombatchai},
  \bibinfo{person}{William~L. Hamilton}, {and} \bibinfo{person}{Jure
  Leskovec}.} \bibinfo{year}{2018}\natexlab{}.
\newblock \showarticletitle{Graph Convolutional Neural Networks for Web-Scale
  Recommender Systems} \emph{(\bibinfo{series}{KDD '18})}.
  \bibinfo{publisher}{Association for Computing Machinery},
  \bibinfo{address}{New York, NY, USA}, \bibinfo{pages}{974–983}.
\newblock
\showISBNx{9781450355520}


\bibitem[You et~al\mbox{.}(2019)]%
        {you2019position}
\bibfield{author}{\bibinfo{person}{Jiaxuan You}, \bibinfo{person}{Rex Ying},
  {and} \bibinfo{person}{Jure Leskovec}.} \bibinfo{year}{2019}\natexlab{}.
\newblock \showarticletitle{Position-aware graph neural networks}. In
  \bibinfo{booktitle}{\emph{ICML}}. PMLR, \bibinfo{pages}{7134--7143}.
\newblock


\bibitem[You et~al\mbox{.}(2020a)]%
        {you2020graph}
\bibfield{author}{\bibinfo{person}{Yuning You}, \bibinfo{person}{Tianlong
  Chen}, \bibinfo{person}{Yongduo Sui}, \bibinfo{person}{Ting Chen},
  \bibinfo{person}{Zhangyang Wang}, {and} \bibinfo{person}{Yang Shen}.}
  \bibinfo{year}{2020}\natexlab{a}.
\newblock \showarticletitle{Graph contrastive learning with augmentations}.
\newblock \bibinfo{journal}{\emph{NeurIPS}}  \bibinfo{volume}{33}
  (\bibinfo{year}{2020}), \bibinfo{pages}{5812--5823}.
\newblock


\bibitem[You et~al\mbox{.}(2020b)]%
        {you2020does}
\bibfield{author}{\bibinfo{person}{Yuning You}, \bibinfo{person}{Tianlong
  Chen}, \bibinfo{person}{Zhangyang Wang}, {and} \bibinfo{person}{Yang Shen}.}
  \bibinfo{year}{2020}\natexlab{b}.
\newblock \showarticletitle{When does self-supervision help graph convolutional
  networks?}. In \bibinfo{booktitle}{\emph{ICML}}. PMLR,
  \bibinfo{pages}{10871--10880}.
\newblock


\bibitem[Zhang and Chen(2018)]%
        {zhang2018link}
\bibfield{author}{\bibinfo{person}{Muhan Zhang} {and} \bibinfo{person}{Yixin
  Chen}.} \bibinfo{year}{2018}\natexlab{}.
\newblock \showarticletitle{Link prediction based on graph neural networks}.
\newblock \bibinfo{journal}{\emph{NeurIPS}}  \bibinfo{volume}{31}
  (\bibinfo{year}{2018}).
\newblock


\bibitem[Zou et~al\mbox{.}(2019)]%
        {zou2019layer}
\bibfield{author}{\bibinfo{person}{Difan Zou}, \bibinfo{person}{Ziniu Hu},
  \bibinfo{person}{Yewen Wang}, \bibinfo{person}{Song Jiang},
  \bibinfo{person}{Yizhou Sun}, {and} \bibinfo{person}{Quanquan Gu}.}
  \bibinfo{year}{2019}\natexlab{}.
\newblock \showarticletitle{Layer-dependent importance sampling for training
  deep and large graph convolutional networks}.
\newblock \bibinfo{journal}{\emph{NeurIPS}}  \bibinfo{volume}{32}
  (\bibinfo{year}{2019}).
\newblock


\end{thebibliography}

%%%%%%%%%%%%%%%%%%%%%%%%%%%%%%%%%%%%%%%%%%%%%%%%%%%%%%%%%%%%%%%%%%%%%%%%%%%%%%%
%%%%%%%%%%%%%%%%%%%%%%%%%%%%%%%%%%%%%%%%%%%%%%%%%%%%%%%%%%%%%%%%%%%%%%%%%%%%%%%
% APPENDIX
%%%%%%%%%%%%%%%%%%%%%%%%%%%%%%%%%%%%%%%%%%%%%%%%%%%%%%%%%%%%%%%%%%%%%%%%%%%%%%%
%%%%%%%%%%%%%%%%%%%%%%%%%%%%%%%%%%%%%%%%%%%%%%%%%%%%%%%%%%%%%%%%%%%%%%%%%%%%%%%
\newpage
\appendix
\section*{Appendix}
\renewcommand\thesubsection{\Alph{subsection}}
\renewcommand\thesubsubsection{\thesubsection.\arabic{subsection}}

\subsection{Reachability Function}

\label{sec:appendix_reachability}

Reachability function $\reach{i}{j}{t}$ is the probability of getting from node $v_i$ to node $v_j$ in $t$ steps. Specifically, in a $t$-step random walk, we start from an initial node $v_i$, and jump to a neighboring node $v_u\in \mathcal{N}_v$ with transition probability $\nicefrac{1}{|\mathcal{N}_v|}$. From $v_u$, the process continues iteratively in the same manner till $t$ jumps. 

Instead of estimating this probability by sampling \cite{sunil2021graphreach}, we calculate it explicitly to keep accurate. It refers to traversing the graph from $v_i$ to $v_j$ according to a probability matrix $\mP=\mD^{-1}\mathcal{E}=[P_{ij}]$, where $\mD\in\mathbb{R}^{|\mathcal{V}|\times|\mathcal{V}|}$ is the diagonal degree matrix with $\mD_{ii}=\sum_{i}\mathcal{E}_{ii}$. Therefore, the reachability function can be formulated as follows:
\begin{equation*}
    \reach{i}{j}{t} = \begin{cases}
        P_{ij} & \mbox{ if $t = 1$ } \\
        \sum_{h}\reach{i}{h}{t - 1} \reach{h}{j}{1}  & \mbox{ if $t > 1$}.
    \end{cases}
\end{equation*}

It can be seen that this is a matrix multiplication operation. To be specific, $\reach{i}{j}{t}$ is the $i$-th row and $j$-th column element of the matrix $\mP^t$, which could be calculated by sparse matrix multiplication. For efficiency, we compute $\mP^t$ for $1\leq t \leq T$ for each dataset and store these $T$ matrices for the position embedding and the $k$-NN similarity prediction task.

\stitle{Complexity analysis.} We first estimate the number of non-zero values of $\mP^t$. Suppose the average node degree is $\delta$. It's easy to see that $\mP$ has about $\delta$ non-zero values per row. Considering the multiplication of $\mP^{t}$ and $\mP$ ($t\geq 1$), each element of $\mP^t$ in a row contributes on average to $\delta$ elements of the new matrix in the same row. Therefore, the average number of non-zero values of $\mP^t$ in a row is at most $R(t)=\min\{\delta^t, |\mathcal{V}|\}$.

Next, we analyze the computation complexity of $\mP^t$ based on sparse matrix multiplication ($1\leq t \leq T$). Considering the multiplication of $\mP^{t}$ and $\mP$, for $i$-th element in each row, we traverse all non-zero values in the $i$-th row of $\mP$ ($\delta$ non-zero values on average), involving a complexity of $O(R(t)\cdot\delta)$ for each row, and $O(R(t)\cdot\delta\cdot |\mathcal{V}|)$ for the matrix multiplication. Thus, the total complexity for reachability calculation is:
\begin{equation*}
    O\left(
        \sum_{t=1}^T R(t)\cdot \delta\cdot |\mathcal{V}|
    \right)=O\left(
        \sum_{t=1}^T R(t)\cdot |\mathcal{E}|
    \right),
\end{equation*}
where $|\mathcal{E}|$ is the number of edges. When $\delta^T\leq |\mathcal{V}|$, the total complexity is $O( \nicefrac{|\mathcal{E}|\cdot \delta\cdot (\delta^T-1)}{(\delta - 1)})=O(|\mathcal{E}|\cdot \delta^T)$. When $T$ is large, since $R(t)\leq |\mathcal{V}|$, an upper bound of complexity is $O(T\cdot |\mathcal{E}| \cdot |\mathcal{V}|)$.

% \subsection{Algorithm and Complexity Analysis}

\label{sec:appendix_algorithm}

% algorithm
% reachability cal
% complexity

\subsection{Fine-tuning Stage of \model}

\label{sec:appendix_ft}

Algorithm \ref{algo:fine-tuning} illustrates the fine-tuning framework. In line 1, we initialize the parameters to an empty set. In line 2-13, we perform prompt-based transferability test to find the most relevant pre-training task for the downstream task. Particularly, in line 2-3, we initialize the parameters with pre-trained parameters, in which $p^{\text{task}(j)}$ is selected as the task embedding initialization. In line 5-9, we fine-tune parameters on the training dataset in the same pipeline as the pre-training. In line 10-13, we test the performance on the validation dataset, and save the parameters if they are the best.

\begin{algorithm}[t]
    \caption{\textsc{Fine-tuning Framework of \model}}
    \label{algo:fine-tuning}
    \LinesNumbered
    \KwIn{Input graph $G=(\mathcal{V}, \mathcal{E}, \mX)$}
    \KwOut{$\mathcal{E'}$}
    $G_1, G_2, \cdots, G_n\leftarrow \text{Connected-components}(G)$\;
    \For{$i=1$ \KwTo $n$}{
        $(\mathcal{V}_i, \mathcal{E}_i, \mX_i)\leftarrow G_i$\;
        $T_i \leftarrow \text{KD-Tree}(\mX_i)$\;
    }
    Construct a complete graph $G'$, which has $n$ nodes, and the weight

    $\theta^*\leftarrow \varnothing$\;
    \For{$j=1$ \KwTo $\left|\theta^{\text{task}}\right|$}{
        Initialize $\theta'$ with $\theta$\;
        Initialize $p^{\text{task}}$ with $p^{\text{task}(j)}$\;
        \Repeat{model convergences}{
            $G'\leftarrow \textsc{Sub-graph}(\mathcal{V}^{\text{train}}, G)$\;
            $G'\leftarrow         \textsc{Prompting}\left(
                G', \mathcal{V}^{\text{train}}, p^{\text{task}}
            \right)$\;
            Fine-tune $\theta'$ and $p^{\text{task}}$ on $G'$\;
        }
        $G'\leftarrow \textsc{Sub-graph}(\mathcal{V}^{\text{val}}, G)$\;
        $G'\leftarrow         \textsc{Prompting}\left(
            G', \mathcal{V}^{\text{val}}, p^{\text{task}}
        \right)$\;
        \If{$\theta^* = \varnothing$ {\bf or} $\theta'$ performs better than $\theta^*$ on $G'$}{
            $\theta^*\leftarrow \theta'$\;
        }
    }
    \Return{$\theta^*$}
\end{algorithm}

\subsection{Further Details of Datasets}

\label{sec:appdendix_datasets}

Table \ref{tab:datasets} shows the details of the five datasets we used. Following \cite{hu2020gpt}, we use 70\% of nodes for pre-training, 10\% for training, 10\% for validation, and 10\% for testing. For all the pre-trained models, we carry out $K$-shot fine-tuning ($K\in\{8, 32, 128\}$) to vary training data size, in which the training set contains $K$ samples for each class sampled from the whole training set, and the validation set is sampled in the same way but from the whole validation set.

\begin{table}[t]
    \caption{Dataset statistics}
    \label{tab:datasets}
    \begin{tabular}{lrrrr}
    \toprule[1pt]
    Datasets & \#Nodes  & \#Edges & \#Feature & \#Labels \\ \midrule
    DBLP & 17,716 & 105,734 & 1,639 & 4 \\
    Pubmed & 19,717 & 88,648 & 500 & 3 \\
    CoraFull & 19,793 & 126,842 & 8,710 & 70 \\
    Coauthor-CS & 18,333 & 163,788 & 6,805 & 15 \\
    ogbn-arxiv & 169,343 & 1,166,243 & 128 & 40 \\
    \bottomrule[1pt]
    \end{tabular}
\end{table}

\subsection{Training Details}

\label{sec:appendix_training_details}

In the pre-training stage, following \cite{hu2020gpt}, we optimize the model via the AdamW optimizer \cite{loshchilov2018decoupled} for at most 500 epochs, and stop training if the validation loss does not drop within 50 epochs. The one with the lowest validation loss is selected as the pre-trained model. 

In the fine-tuning stage, we follow GPPT framework \cite{sun2022gppt} introduced in \sref{sec:downstream} to optimize the pre-trained models for at most 500 epochs, in which we train prototype nodes for each class and maximize the similarity between query nodes and corresponding prototype nodes on the $K$-shot training set. The optimization setting remains the same as the pre-training stage. We save the models with the lowest validation loss on the $K$-shot validation set. 

For stability, we use five different fixed seeds to sample data from the training set and validation set for each $K$-shot setting, and record the average performance tested on the whole test set as a result. We repeat this process five times, and report the average and the standard deviation of these results. Therefore, the reported standard deviation results from model initialization and optimization, rather than the data few-shot sampling.

For the hyper-parameters, we set $t=9$, $t'=6$, $\alpha=0.5$, $\alpha'=1$, batch size as 256, and learning rates as 0.001 for both the pre-training stage and the fine-tuning stage. The number of anchors $m$ is selected as $0.01 \times |\mathcal{V}|$. $w^{\text{pos}}$ is selected from $\{0.01, 0.1, 1\}$. Finally, we set $w^{\text{pos}}=1$ for DBLP and CoraFull, $w^{\text{pos}}=0.1$ for Coauthor-CS, and $w^{\text{pos}}=0.01$ for Pubmed and ogbn-arxiv. For GNN backbones, we set the hidden dimension as 64 on DBLP, Pubmed, CoraFull, Coauthor-CS, and 128 on ogbn-arxiv. The number of GNN layers is 3, and the head number is 8 for all datasets.

\end{document}